\documentclass[10pt]{article}
\usepackage{isolatin1}
\usepackage{multirow}
\usepackage{calc}
\usepackage{epsfig,amsfonts,amsthm,latexsym,amsmath}

\usepackage{fleqn}
\usepackage{tikz}
\usepackage{lscape}
\usepackage{xcolor}
\usepackage{pslatex}
\usepackage{hyperref}
\usepackage{colortbl}
\usepackage{booktabs}
\usepackage{multirow}
\usepackage{amssymb}
\usepackage{natbib}
\usepackage{graphicx}

\title{\normalsize\bf%
THE MULTI-VEHICLE COVERING TOUR PROBLEM: BUILDING ROUTES FOR URBAN PATROLLING
}

\author{Washington A. Oliveira$^{1*}$, \ Antonio C. Moretti$^1$ \ and \ Ednei F. Reis$^2$}

\begin{document}

\date{}

\maketitle

\vspace{-20pt}
\begin{center}
{\footnotesize 

*Corresponding author\\
$^1$Faculdade de Ci\^encias Aplicadas, Universidade Estadual de Campinas, 13484-350 Limeira, SP, Brasil\\
$^2$Departamento de Matem\'atica, Universidade Tecnol\'ogica Federal do Paran\'a, 84016-210 Ponta Grossa, PR, Brasil \\
E-mails: washington.oliveira@fca.unicamp.br [Oliveira]/moretti@ime.unicamp.br [Moretti]/edneif@utfpr.edu.br [Reis]
}\end{center}

\bigskip
\noindent
{\small{\bf ABSTRACT.}
In this paper we study a particular aspect of the urban community policing: routine patrol route planning. We seek routes that
guarantee visibility, as this has a sizable impact on the community perceived safety, allowing quick emergency
responses and providing surveillance of selected sites (e.g., hospitals, schools). The planning is restricted to the
availability of vehicles and strives to achieve balanced routes. We study an adaptation of the model for 
the multi-vehicle covering tour problem, in which a set of locations must be visited, whereas another subset must be close enough to the
planned routes. It constitutes an NP-complete integer programming problem. Suboptimal solutions are
obtained with several heuristics, some adapted from the literature and others developed by us. We solve some adapted instances from TSPLIB and an instance with real data, the former being compared with results from literature, and latter being compared with empirical data.
}


\baselineskip=\normalbaselineskip

\section{Introduction}

{\it Community policing} aims to serve several objectives: to gather information related to the community needs, to prevent crimes, to quickly respond to emergencies, to monitor public buildings, etc. The two most common ways of organizing patrol car operations are the allocation of a car to a certain fixed geographical location and the allocation of a car to a certain route covering a larger area. The second is the method usually chosen by the municipal guards and military police in the state of S\~ao Paulo, Brazil. The planning of these routes rarely adopts a scientific approach, empirical rules are used instead.

One of the most important features of community policing is the contact between the members of the community and the 
patrolling officers, improving the exchange of information between the community and the police force. 
Balanced routes with respect to the number of visits can improve the whole process of community patrolling, since this can reduce the number of persons 
and visits per officer.

Using a fleet of known size, we consider how to efficiently construct routes in order to patrol a given geographical area. The routes should 
satisfy the following criteria: a certain set of sites must be included in the routes; a second set may or may not belong to the routes; a third set is made
up of sites that must be observed (\textit{covered}) by the patrolling officer, in the sense that these sites are not visited, but they must be close enough to at least one visited site; the 
number of routes must equal the number of available vehicles; the routes must be balanced, that is, the number of visited sites for 
each route is approximately the same; all routes must start and finish at the same geographical point, the \textit{base} of operations.

The problem is modeled as an integer programming problem. The model presented in this paper is related to the~\textit{multi-vehicle Covering Tour Problem} ($m$-CTP), where we discard the vehicle capacity constraints and include a certain balance among the vehicles. The resulting combinatorial optimization problem is NP-complete, which justifies the use of heuristics developed in this paper to obtain suboptimal solutions of acceptable quality in reasonable time.

The heuristics were implemented in $\textsc{Matlab}^{\scriptsize{\textregistered}}$ and compared using several instances from TSPLIB. These instances, together with real data from the city of Vinhedo, S\~ao Paulo, Brazil, were used to validate the code.

In Section~\ref{sec:Scope}, we define the balanced multi-vehicle urban patrolling problem. The mathematical formulation 
is described in Section~\ref{sec:Formulation}. The heuristics developed in this paper are presented in detail in Section~\ref{sec:heuristics}. In 
Section~\ref{sec:Numerical}, we present the numerical experiments. Finally, the conclusions are given in Section~\ref{sec:conclusions}.

\section{Building routes for urban patrolling}\label{sec:Scope}

In the problem considered in this paper, a set of geographical points that need to be 
visited during a routine patrol was determined by the police force, which might include schools, hospitals, public buildings, etc. Additionally, there is 
another set that must be at convenient distance from the route, for instance, public parks, community centers, bank agencies, etc. 
Good designs of routes are of crucial importance, due to the limited resources (the size of the fleet) available to cover usually 
large geographical areas. Therefore, the patrolling officers must be guided from one visit to another in order to avoid bad 
empirical circuits.

More specifically, we will build routes for urban patrolling for the city of Vinhedo. This city is located approximately 80 km east of the city of S\~ao Paulo, Brazil. Its economy is mostly 
based on agriculture, in particular the growing of grape and production of related goods, specially wine. It occupies an area measuring 
roughly 81.742 km$^2$, and, as of 2010, had about 63,500 inhabitants. It is part of a group of cities known as the \textit{fruit circuit}, which 
promotes three to four festivals per year, the fig festival, the strawberry festival, for example. The per capita income is high and 
the criminality rate is low.

The Municipal Guard of the city of Vinhedo must assist the population in public safety matters, crime prevention being the main objective to be achieved 
by the police force. Preventive measures that increase the contact between the patrolling officers and members of the community are of fundamental importance 
to community policing~\cite[]{CommunityPolicing}. This strategy aims to build a partnership between the community and Municipal Guard based on the premise that a collective effort must be made to improve public 
safety. This means that the people in a certain area have to not only participate in the discussions about safety and establish priorites and strategies, 
but also share with the officers that patrol this area the responsibility for the safety of the region. We can highlight the following measures: organization of public audiences to discuss 
the community problems in order to develop strategies and priorites; mobilization of the community for self-protection and solving problems that generate 
crimes; motivate the frequent dialogue between the officers and the community.

Ideally, the patrolling officers must stop in each visit point, interact with the people there, and watch certain points of interest within viewing 
distance. Aiming a greater proximity with the community, the chief of operations needs to designate a balanced set of visits for the patrolling 
cars, reducing the number of persons per officer. However, in practice, the city is divided into sectors (see Figure~\ref{fig:DistributionVinhedo}), and a certain number of vehicles is assigned 
to a given sector. Moreover, the patrolling officers make the circuit of visits empirically, based on their knowledge 
of the geographical area, which may be unproductive. Figure~\ref{fig:DistributionVinhedo} illustrates the distribution of five sectors within the Vinhedo 
city region, denoted by A, B, C, D and E. Note that the geographical points are not equally distributed within these sectors.

There are regions where the geographical points of interest are more concentrated than others, due to the characteristics of a geographical 
area (e.g., the distances between two schools may be much larger in peripheral areas than in central ones). Consequently, if we obtain balanced routes, there 
might be shorter routes in the central regions than in peripheral ones. Therefore, the fact that there are routes which are shorter than others is not inconsistent with our proposal for building routes for urban 
patrolling, since the efficiency of the contacts established by the patrolling officers with the members of the community outweighs the balance among 
the lengths of routes.

In our approach, we aim to build routes in the context urban patrolling that balance the number of visits without a prior division of the 
geographical area into sectors. The set of visits for each patrolling car is decided by dividing the set of all visits into clusters. Once the set of visits 
for a patrolling car is obtained, to determine the order to the visits, we have chosen to minimize the overall length of all routes instead of choosing the order empirically. 
Thus, the problem is modeled as an integer programming problem whose the main objective is to determine balanced circuits in number of visits for each patrolling car.

A proper survey of the features of the problem was conducted with Vinhedo patrol professionals. Ideally the distances (or travel times) among geographical points would be calculated separately from real data, taking into account possible 
paths between nodes, average speeds, etc. This would imply in non-symmetrical distances. Nevertheless, in the real instance considered, involving the city of Vinhedo, the true path lengths (corresponding to routes using the city streets) 
were not known. Therefore, the Euclidean distance was chosen as an approximation to determine each route. Note that once the order of visits is established, the 
patrolling officers can use their experience to decide how to go from one visit to the next.

There are similarities between the design of routes for urban patrolling and the model presented in this paper, namely, the construction of circuits covering certain geographical points, while visiting mandatory and optional ones. The number of visits is balanced and the number of routes is fixed 
(we do not want cars and personnel idling at the base). Throughout the text we present our model, and more importantly, our strategy 
to obtain approximate solutions.

The model involves a graph $G=(V\cup W, E)$, whose nodes correspond to strategic geographical points, e.g., intersections, certain locations, 
etc., that are either important on its own, or serve to establish reference locations for the routes. 
The node set is partitioned into two subsets: $V=\{0,\ldots,n\}$ is the set of nodes that may belong to routes and $W=\{n+1,\ldots,n+\ell\}$ 
is the set of nodes that must be covered, but not visited. The set $V$ contains a subset $T$ of nodes that must be visited. 
Node $0 \in T$ corresponds to the base. The set $V\setminus \{0\}$ is denoted by $V^*$. The symbol $T^*$ denotes the set $T\setminus \{0\}$. 

The set $E$ contains all possible undirected 
arcs between nodes of $V$. The entry $c_{ij}$, of the distance matrix $C=(c_{ij})$, contains the Euclidean distance between nodes 
$i,j = 0, 1,\ldots,n+\ell$. This assumption implies that $C$ is symmetric with zero diagonal. In this case, we choose 
to represent the set (undirected arc) $\{i,j\}$ by the ordered pair $(i^*, j^*)$, where $i^*=\min\{i,j\}$ and $j^*= \max\{i,j\}$. The size of the fleet is denoted by $m$ and the admissible distance from a node in the route to a node that must be 
covered is denoted by $c$. We need one last parameter, to express our tolerance regarding the lack of balance between different routes. We compare 
routes by means of the number of nodes each route contains. The number $r$ denote the maximum difference allowed for the total visited 
nodes in any two routes.

We want to construct $m$ routes satisfying the following conditions:
\begin{itemize}
 \item Each route is a circuit in $G$ containing node 0.
 \item Each node in $V^*$ must belong to at most one route.
 \item Each node in $T^*$ must belong to exactly one route.
 \item There must be at least a visited node at a distance of at most $c$ from each node in $W$.
 \item The number of nodes in any two routes differ by at most $r$.
\end{itemize}

Variants of this model have been used in many contexts.~\cite{Art:labbe1986maximizing} discuss how to simultaneously decide on the location of 
mailboxes and the planning of routes involving candidate sites, such that all users are close enough to some mailbox. The planning of routes 
for medical mobile units~\cite[]{Art:brown1995us,Art:foord1995gambia,Art:Hodgson:JORS113,Art:oppong1994spatial,Art:SWADDIWUDHIPONG01061995} 
bears many similarities with the planning of urban patrol routes. In places where medical services for small villages are rendered by mobile units, 
their routes must be planned in such a way that, in addition to visiting specific locations, the villages not included in the tour must be within 
walking distance from some other village included in the tour. The model can also fit situations in business sectors, 
see~\cite{Art:simms1989fixed} for an application in dairy practice. 

The special case of $m$-CTP where $V=T$ reduces to 
a vehicle routing problem~\cite[]{Art:Laporte1992345} with unit demand.~\cite{Art:Baldacci} present three Scatter Search 
heuristic algorithms for the 1-CTP.~\cite{motta2001grasp} use a GRASP approach to solve a variant of the 1-CTP, the 
Generalized Covering Tour Problem, whose minimum length tour may pass through a subset of $W$. In another point of view,~\cite{jozefowiez2004multi,jozefowiez2007bi} present a multi-objective covering tour problem, a generalization of the 
1-CTP where the parameter $c$ is omitted and replaced by an objective. They propose a hybrid strategy approach that combines a 
multi-objective evolutionary algorithm with a branch-and-cut algorithm to determine the Optimal Pareto sets. 

~\cite{Art:Ha} obtained exact solutions for a variation of $m$-CTP, where the constraint on the length of the routes is relaxed. They used a branch-and-cut algorithm and a metaheuristic (based on the evolutionary local search) to obtain those solutions. ~\cite{Art:Jozefowiez} and~\cite{Art:Lopes} used a branch-and-price algorithm, in which a column generation approach is applied at each node of the search. Despite the recent applications for $m$-CTP using variations of the exact methods branch-and-bound, branch-and-cut, and branch-and-price, we did not succeed in solving large instances. In our application, in the context of urban patrolling, the instances are of the order of more than two thousand points.

\section{Mathematical formulation of the problem}\label{sec:Formulation}

In order to facilitate the modeling of the covering restrictions, we define the set of nodes in $V$ within the allowed prescribed 
distance $c$ from each node $j\in W$: $S_j = \{i\in V\ |\ c_{ij}\leq c \}$. We may suppose without loss of generality that $|S_j|\geq 2$, since if there is only one node 
$i\in V$ close enough to some $j$, then we may as well include $i\in T$ and eliminate $j$ from $W$. Similarly, we may assume that 
$S_j \cap T = \emptyset$, for all $j$, since we do not need to worry about covering nodes that are close enough to some node in $T$. 
With these assumptions, $S_j = \{i\in V\setminus T\ |\ c_{ij}\leq c\}$ and its cardinality is at least two, for all $j\in W$.

The model contains two sets of binary variables. The variable $y_{ik}$ is 1 if node $i$ is visited by vehicle $k$, and 0 otherwise, for $i\in V$, 
$k=1,\ldots,m$. The variable $x_{ijk}$ is 1 if vehicle $k$ uses arc $(i,j)$ in its route, and 0 otherwise.

\begin{eqnarray}
\mbox{Min}&& \sum\limits_{k=1}^m\sum\limits_{i=0}^{n-1}\sum\limits_{j=i+1}^{n} c_{ij}x_{ijk}\ , \label{eq:FO1}\\
\mbox{s.t.}\ && \sum\limits_{k=1}^m\sum\limits_{i\in S_j} y_{ik}\geq 1, \quad j\in W, \label{eq:RE2}\\
&& \sum\limits_{k=1}^m y_{ik}\leq 1, \quad i\in V\setminus T, \label{eq:RE3}\\
&& \sum\limits_{k=1}^{m} y_{ik}=1, \quad i\in T^*,  \label{eq:RE8}\\
&& \sum\limits_{i=0}^{h-1} x_{ihk} + \sum\limits_{j=h+1}^{n} x_{hjk} = 2y_{hk},\quad h\in V^*,\quad k=1,\ldots,m, \label{eq:RE4}\\
&& \sum\limits_{{\scriptsize \begin{array}{l}i\in S,\ j \in V\setminus S \\ \mbox{or}\  j\in S,\ i \in V\setminus S \end{array}}} x_{ijk} \geq 2 y_{hk}, \quad \begin{array}{l} \\ S\subset V^*,\ h\in S,\\ 2\leq |S|\leq n-1,\ k=1,\ldots,m,\end{array} \label{eq:RE5}\\
&& \sum\limits_{i=1}^{n} x_{0ik}=2, \quad k=1,\ldots,m,  \label{eq:RE6} \\
&& \sum\limits_{i=1}^{n} y_{ip} - \sum\limits_{i=1}^{n} y_{iq} \leq r, \quad p,q=1,\ldots,m,  \label{eq:RE7}\\
&& \sum\limits_{i=1}^{n} y_{ip} - \sum\limits_{i=1}^{n} y_{iq} \geq -r, \quad p,q=1,\ldots,m,  \label{eq:RE7-2}\\
&& y_{0k}=1, \quad k=1,\ldots,m,  \label{eq:RE9}\\
&& y_{ik}, x_{ijk} \in \{0,1\}, \quad i,j \in V, \quad k=1,\ldots,m.  \label{eq:RE10}
\end{eqnarray}

Using these variables, the ``cost" of a solution, given in~(\ref{eq:FO1}), is the cumulative length of all routes, which we wish to minimize. The 
constraints are modelled as follows. The covering of node $j\in W$ is guaranteed by~(\ref{eq:RE2}). 
Constraints~(\ref{eq:RE3}) make sure that each node in $V\setminus T$ belongs to at most one route. The fact that every node in $T$ must be visited by some tour is 
expressed in~(\ref{eq:RE8}). Constraints~(\ref{eq:RE4}) imply that, if node $h$ belongs to 
route $k$, then it has two neighbors in the route. Constraints~(\ref{eq:RE5}) avoid subtours, by forcing that, if node $h\in S \subseteq V^*$ belongs to route $k$, then the cut-set $(S,V\setminus S)$ 
must contain at least two arcs of route $k$.~(\ref{eq:RE6}) guarantees that each route has two arcs incident to the base. The maximum 
difference between the number of nodes of different routes is enforced by~(\ref{eq:RE7}) and~(\ref{eq:RE7-2}), i.e., the maximum difference 
$\tilde{r}=\max\limits_{1\leq p,q \leq m}\left|\sum\limits_{i=1}^{n} y_{ip} - \sum\limits_{i=1}^{n} y_{iq}\right|$ obtained in the solution 
must be less than or equal to the parameter $r$. Constraints~(\ref{eq:RE9}) force that the base belongs to every route. The last set of constraints,~(\ref{eq:RE10}), simply specifies the allowed values for the variables.

\section{Heuristics}\label{sec:heuristics}

The heuristics developed for the balanced multi-vehicle urban patrolling problem modeled in the last section are divided into three phases. In Phase 1, subsets $V_k$ and $W_k$, $k=1,\ldots,m$, 
are selected, where $V_k$ and $W_k$ are the nodes that may be visited and that should be 
covered by route $k$, respectively. Phase 2 deals with $m$ 1-CTP problems, defined on the subgraph 
induced by $V_k \cup W_k$. At this point, $m$ closely related problems are considered separately. 
The last phase tries to improve the solution by taking this interrelation into account. The three-phase sequence is repeated 
according to a criterion specific to the routine employed in Phase 1. The best solution in the whole 
loop of three phases (or \textit{outer iterations}) is selected.

Table~\ref{tab:RoutePhase} below summarizes the routines employed in each phase of the various heuristics. The routine employed in Phase 1 will lend 
its name to the heuristic. Note that they share Phase 2, and the first three heuristics use Balanced 2-opt in Phase 3, while the Sector Partition uses 
Multicover Elimination.

\begin{table}[!htb]
\begin{center}
\begin{tabular}{lclcl}
\toprule
Phase 1 & & Phase 2 & & Phase 3 \\ \toprule
Greedy & & & &  \\ 
Selection  & & & &  \\ \cmidrule{1-1}
Sweep & &  Modified  & & Balanced \\ 
Routine & &1-CTP & &2-opt \\ \cline{1-1}
Route-first/ & &  Routine & &  \\ 
Cluster-second & &  & &  \\ \cmidrule{1-1}\cmidrule{5-5}
Sector & &  & & Multicover \\ 
 Partition & &  & &  Elimination \\ \toprule 
\end{tabular}\caption{Routines according to Phase}\label{tab:RoutePhase}
\end{center}
\end{table} 

Our mathematical model resembles the model in~\cite{Art:hachicha2000heuristics}. However, in that model the length and number of visits per tour are limited and the number of routes is variable, whereas in this model the number of visits is balanced and the number of routes is fixed. In order to take into account these differences, we consider modified versions of the sweeping algorithm, Route-first/Cluster-second algorithm and 2-opt$^*$ algorithm presented in their paper. The \textit{Sweep Routine} corresponds to steps 1 and 2 of the sweeping algorithm. Similarly, \textit{Route-first/Cluster-second} is formed by steps 1 and 2 of the algorithm of same name. The \textit{Balanced 2-opt} routine contemplates improvements via arc swapping, and is adapted from the 2-opt$^*$ algorithm.

The \textit{Modified 1-CTP Routine} is a modification of the heuristic described in~\cite{Art:gendreau1997covering} for the covering tour problem. The remaining routines were developed by us.

The \textit{Greedy Selection} routine gradually selects sites using a criterion that selects the nearest site to the one previously selected, forming a circular ordered list. 
The nodes in this list are then partitioned into $m$ subsets of approximately equal size, keeping the order of the original list and 
starting with the first node in the list. In subsequent iterations, the selection step is not repeated. Instead, we simply shift by one the 
order of the nodes in the list and redo the partitioning. This is repeated approximately $t/m$ times, where $t$ is the cardinality of the list.

In the \textit{Sector Partition} routine, we use a geographical approach to divide $V$, $T$, and $W$, into $m$ subsets, corresponding to circular 
sectors. Each outer iteration corresponds to a counterclockwise shift of the sectors. This simple geographical division is used to 
reduce the computational time in Phase 1.

The \textit{Multicover Elimination} checks whether some node in $W$ is covered by more than one node included 
in a route. If this is the case, there may be room for improvement, by removing one of the superfluous nodes.

In the described routines, it is often necessary to know the subset of nodes in $W$ covered by a particular node in $V$. The subset 
covered by node $i$ is denoted by $C_i = \{ j\in W\ |\ i\in S_j\}$. Recall that $S_j=\{\ell\in V\setminus T\ |\ c_{\ell j}\leq c\}$.

\subsection{Greedy Selection}

Initially we form a single route $R=(h_0,h_1,\ldots,h_z)$ that contains all nodes in $T$ and covers all nodes in $W$ as follows. The routine starts 
with $h_0=0$, $R=(h_0)$ and $L=T^{*} \cup W$. The set $L$ is gradually emptied using the criterion that selects the nearest site to the one 
previously selected in $L$. If the chosen node $h$ belongs to $T$, it is simply appended to $R$, and $\{h\}\cup C_h$ is removed from $L$. 
If $h$ belongs to $W$, then one selects from $S_h$ the node that covers the greatest number of yet uncovered nodes, say node $\ell$, and 
appends it to $R$. Then $C_{\ell}$ is removed from $L$.

In the next two paragraphs, we detail the main differences between our approach and Hachicha's Sweep and Route-first/Cluster-second routines. There, the 
selection of the sets $V_k$ and $W_k$, $k=1,\ldots,m$, is made in order to minimize the number of routes, while satisfying the demand, the capacity of the 
vehicle and their constraints on the length of the routes. In our approach, we choose $V_k$ and $W_k$ with approximately the same number of elements to keep the balance among 
routes.

Here, and in the next two routines, once $R$ is constructed we consider the sequence $R^*=R\setminus \{0\}$ as a circular list. In order to apply Phase 2, 
we need to divide $V$, $T$, and $W$ 
into $m$ subsets that are induced by a partition of $R^*$ as follows. Let $p=\lfloor z/m \rfloor$, $q=z-mp$, where $z$ denotes 
the number of nodes in $R^*$. Starting from 
the first node in $R^*$, we select $q$ subsets of sequential nodes of size $p+1$, and $m-q$ subsets of size $p$. The nodes 
in $V$ (resp., $T$) in the $k$th subset union $\{0\}$ are denoted by $V_k$ (resp., $T_k$). The $k$th subset of $W$ is $W_k=\bigcup \{C_i\ |\ i\in V_k\}$.

In the next round, the partition of $R^*$ will start at the second node, and then third, and so on, in a total of $p$ or $p+1$ outer iterations, 
depending on whether $z$ is a multiple of $m$ or not.

\subsection{Sweep Routine}

The difference between this routine and the previous one is the strategy used to build the single route 
$R=(h_0,h_1,\ldots,h_z)$. The sweeping process is applied to the 
vertices $T\cup W$ and several solutions are generated. The routine starts with $h_0=0$, $R=(h_0)$ and $L=T^{*} \cup W$. Choose an arbitrary 
node $\bar{h}\in L$ and consider a half line from $h_0$ passing through $\bar{h}$. The set $L$ is gradually emptied using 
the criterion that sweeps the nodes $h \in L$ according to the ascending order of the angles $\theta_h = \widehat{\bar{h} h_0 h}$. If 
the chosen node $h$ belongs to $T$, it is simply appended to $R$, and $\{h\}\cup C_h$ is removed from $L$. 
If $h$ belongs to $W$, then one selects from $S_h$ the node that covers the greatest number of yet uncovered nodes, say node $\ell$, and 
appends it to $R$. Then $C_{\ell}$ is removed from $L$. Once $R$ is constructed, the process continues as in the Greedy Selection routine.

\subsection{Route-first/Cluster-second}

Again, the difference between this routine and Greedy Selection routine is the strategy that we build the single route 
$R=(h_0,h_1,\ldots,h_z)$. Here, a feasible 1-CTP solution for $V$, $T$, and $W$ is determined by means of the 
Modified 1-CTP Routine (section~\ref{sec:1CTP}), say route $R=(h_0,h_1,\ldots,h_z)$, $h_0 = 0$, that contains all nodes in $T$ and covers all nodes in $W$. 
This tour is then divided into smaller feasible routes as in the Greedy Selection routine. 

\subsection{Sector Partition}

This routine is applicable only in cases where there is a geographical model of the problem. The site 
associated with the base node is taken as the geographical center of a circular disk containing all locations (nodes) under consideration. 
This disk is partitioned into $m$ circular sectors of same central angle. Nodes corresponding to sites in the $k$th partition form the 
sets $V_k$, $T_k$, and $W_k$. The first partition is arbitrary. In the next iteration, the sectors are shifted counterclockwise by $360\textordmasculine /t$, 
and this is repeated until we return to the original partition, a total of $t$ outer iterations. In the computer experiments, the value $t=10$ was used. 
Due to its simplicity, if the distribution of nodes is non-uniform, this way of choosing the partitions $V_k$, $T_k$, and $W_k$ 
does not guarantee the equilibrium among the number of nodes in each partition, possibly producing unbalanced routes.

\subsection{Modified 1-CTP Routine}\label{sec:1CTP}

\cite{Art:gendreau1997covering} developed a heuristic for the covering tour problem (1-CTP), using elements of 
the GENIUS heuristic for the traveling salesman problem (TSP) of~\cite{Art:gendreau1992new} and PRIMAL1 set covering heuristic 
of~\cite{Art:Balas}.

This heuristic uses the fact that, if the set of nodes that should be visited (that is, the support of the optimal $y$) is known, 
then the covering tour problem reduces to a traveling salesman problem. This suggests the construction of a covering problem by 
considering the ``covering" aspect separately, namely the combinatorial optimization problem~(\ref{eq:RE11}) below. Notice that, since $m=1$, there is no need 
for the tour index $k$.
\begin{eqnarray}
\mbox{Min}&& \sum\limits_{j\in V}c_j y_j\ ,\nonumber\\
\mbox{s.t.}\ && \sum\limits_{j\in S_i} y_j \geq 1, \quad i\in W, \label{eq:RE11}\\
&& y_j=1, \quad j\in T, \nonumber\\
&& y_j \in \{0,1\}, \quad j \in V.  \nonumber
\end{eqnarray}

In PRIMAL1, variables $y_j$ are gradually included in the solution according to a greedy rule, using one of three merit functions: 
(i) $f(c_j,b_j) = c_j/(\log_2 b_j)$; (ii) $f(c_j,b_j) = c_j/b_j$; (iii) $f(c_j,b_j) = c_j$; where $b_j$ is the number of uncovered 
(in the current partial solution) nodes in $W$ that are covered by $j$, and $c_j$ is the cost to insert $j$ in the current solution.

On the other hand, TSP routes are also constructed incrementally in GENIUS. During the construction phase, a tentative tour is built 
starting with three arbitrarily selected nodes and using a general insertion procedure (GENI), with rules for selecting candidates 
and rules for evaluating the inclusion in the tour. Once a complete tour is obtained, one seeks to improve it by removing and reinserting each 
node of the tour, in a post-optimization procedure called US (for Unstringing and Stringing).

GENIUS and PRIMAL1 are combined to produce a covering tour heuristic as follows. Nodes are gradually selected 
and added to the set of nodes that should be visited, and 
a new approximate solution to the TSP involving each insertion of these nodes is obtained using GENIUS. This procedure stops when all nodes 
in $W$ have been covered. The selection of the each node is made by considering the merit functions in PRIMAL1 heuristic using the 
coefficient $c_j$ (cost to insert this node in the 
approximate solution to the TSP, calculated by GENI selection rules). At the end, post-optimization is applied by removing superfluous 
nodes. The whole process is done for each merit function, two sequences of the merit functions (i)-(ii)-(iii) and (i)-(iii)-(ii) are considered, 
and the best solution is selected. 

Numerical experiments with our code for the various heuristics showed that the GENI part was fast and produced acceptable results, whereas 
the post-optimization US routine was quite costly. After several trials, we arrived at a modified routine of Gendreau et al.'s covering tour heuristic, 
in order to avoid the repeated construction of unnecessary approximate solutions of the TSP at the insertion and removal of each node. This is done by inserting the new node using GENI rules  and removing superfluous 
nodes using US rules without discarding the previous TSP solution.

More details about the experiments with the different heuristics tried out, 
which showed a better computational performance with these modifications, can be found in~\cite{Tese:Wash}. In the modified routine listed below, 
the merit functions are used in the sequence (i)-(ii)-(iii). 

\begin{enumerate}
\item[] \noindent\textbf{STEP 1.} Let $H \leftarrow T$, $\bar{z}\leftarrow \infty$. Let $f$ be the merit function specified in (i). Let $z$ be the 
value of the tour obtained applying GENIUS to the TSP involving the nodes in $H$. If all nodes in $W$ are covered, go to Step 2. 
Otherwise, go to Step 3.

\item[] \noindent\textbf{STEP 2.} If $z\leq\bar{z}$, let $\bar{z}\leftarrow z$, $\bar{H}\leftarrow H$. If the definition used for the merit 
function is the last, stop with the best solution so far, with value $\bar{z}$ and set of visited nodes $\bar{H}$. Otherwise, remove from 
$H\setminus T$ nodes associated with multiple covered nodes in $W$ using US. Go to Step 3 with the next definition of the merit function $f$.

\item[] \noindent\textbf{STEP 3.} Select within $V\setminus H$ node $h^*$ using PRIMAL1, with cost coefficients $c_j$ calculated using 
criteria defined in GENI. Insert $h^*$ in the route employing GENI, and let $H\leftarrow H \cup \{h^*\}$. Repeat the process until all nodes in $W$ are covered. Let $z$ be the 
value of the tour obtained. Go to Step 2.
\end{enumerate}

When this routine is used in Phase 2, it is run $m$ times with the sets $V_k$, $T_k$, and $W_k$, for $k= 1,\dots,m$, constructed in Phase 1.

\subsection{Balanced 2-opt}

This is a modification of the 2-opt$^*$ heuristic. The changes introduced aim to keep the balance, 
in terms of number of nodes, among the routes. Steps 1 and 2 are adaptations, since in their approach, the number of the routes can decrease and changes between arcs are allowed provided the length of the routes and number of visits are within the preset values, while in our approach, those changes can be made only if the balance between routes is not destroyed and the number of routes is not altered (we do not want cars and personnel idling at the base). Step 3 is different, it considers swapping of nodes belonging to different routes. Notice that the exchanges considered in this last step do not alter the number of nodes in each route. 

At this point, the initial set of $m$ routes constructed in Phase 1 has already undergone improvement in Phase 2, and the following procedure is executed.

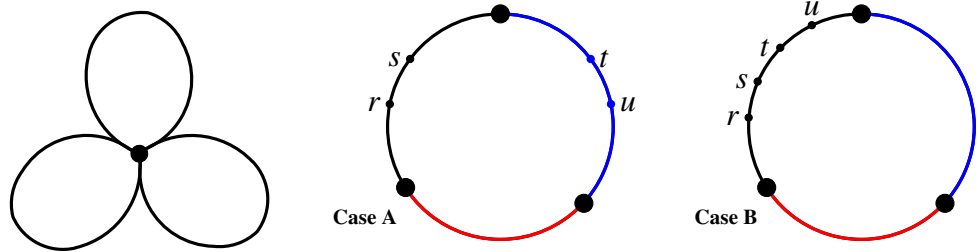
\begin{figure}[t]
\begin{center}

\begin{tikzpicture}[scale=0.6]

{\pgftransformxshift{-6cm}\pgftransformyshift{1.4cm}\pgftransformrotate{-16.5}\pgftransformscale{0.93}
\draw[line width=1.2pt, rotate=90] (0,0) arc (175:90:2cm) -- (2,1.83) arc (90:10:1.3cm) arc (360:310:1.3cm) arc (315:220:1.8cm) -- cycle;
\draw[line width=1.2pt, rotate=210] (0,0) arc (175:90:2cm) -- (2,1.83) arc (90:10:1.3cm) arc (360:310:1.3cm) arc (315:220:1.8cm) -- cycle;
\draw[line width=1.2pt, rotate=330] (0,0) arc (175:90:2cm) -- (2,1.83) arc (90:10:1.3cm) arc (360:310:1.3cm) arc (315:220:1.8cm) -- cycle;

\filldraw[black](0,0) circle(.2);}

\draw [line width=1.2pt] (2,2) circle (2.5cm);
\draw [line width=1.2pt] (10,2) circle (2.5cm);

\draw [blue, line width=1.2pt] (2,2) +(320:2.5cm) arc (320:360:2.5cm);
\draw [blue, line width=1.2pt] (2,2) +(0:2.5cm) arc (0:90:2.5cm);
\draw [red, line width=1.2pt] (2,2) +(210:2.5cm) arc (210:320:2.5cm);
\draw [blue, line width=1.2pt] (10,2) +(320:2.5cm) arc (320:360:2.5cm);
\draw [blue, line width=1.2pt] (10,2) +(0:2.5cm) arc (0:90:2.5cm);
\draw [red, line width=1.2pt] (10,2) +(210:2.5cm) arc (210:320:2.5cm);

\node at (-1,0) {\footnotesize \textbf{Case A}};
\node at (7,0) {\footnotesize \textbf{Case B}};

\path (-0.1,0.65) coordinate (t2);
\path (7.9,0.65) coordinate (t02);

\path (2,4.5) coordinate (t3);
\path (10,4.5) coordinate (t03);

\path (3.85,0.3) coordinate (t4);
\path (11.85,0.3) coordinate (t04);

\path (0,3.5) coordinate (t5);
\path (4,3.5) coordinate (t05);

\path (-0.45,2.5) coordinate (t6);
\path (4.45,2.5) coordinate (t06);

\path (8.9,4.25) coordinate (t7);
\path (8.2,3.75) coordinate (t07);
\path (7.7,3) coordinate (t8);
\path (7.5,2.2) coordinate (t08);

\filldraw[black] (t2) circle(.2);
\filldraw[black] (t02) circle(.2);

\filldraw[black] (t3) circle(.2);
\filldraw[black] (t03) circle(.2);

\filldraw[black] (t4) circle(.2);
\filldraw[black] (t04) circle(.2);

\filldraw[black] (t5) circle(.08);
\filldraw[blue] (t05) circle(.08);

\filldraw[black] (t6) circle(.08);
\filldraw[blue] (t06) circle(.08);

\filldraw[black] (t7) circle(.08);
\filldraw[black] (t07) circle(.08);

\filldraw[black] (t8) circle(.08);
\filldraw[black] (t08) circle(.08);

\draw (t5) node[anchor=east] {\large $s$};
\draw (t05) node[anchor=west] {\large $t$};

\draw (t6) node[anchor=east] {\large $r$};
\draw (t06) node[anchor=west] {\large $u$};

\draw (t7) node[anchor=south] {\large $u$};
\draw (t07) node[anchor=east] {\large $t$};

\draw (t8) node[anchor=east] {\large $s$};
\draw (t08) node[anchor=east] {\large $r$};

\end{tikzpicture}
\caption{How to build a single route for $m=3$, and two patterns of arc-pairs}\label{fig:replicadeposito_opt}
\end{center}
\end{figure}

\begin{figure}[h]
\begin{center}
\begin{tikzpicture}[scale=0.8]

\draw[line width=1.2pt] (4,7) arc (360:180:2cm);
\draw[line width=1.2pt] (4,7) arc (0:15:2cm);
\draw[line width=1.2pt] (0,7) arc (180:165:2cm);

\draw [line width=1.2pt] (2,7) +(45:2cm) arc (45:135:2cm);
\draw [line width=1.2pt, dashed] (0.55,8.4)--(3.5,8.4);
\draw [blue, line width=1.2pt, dashed] (0.05,7.5)--(3.95,7.5);
\draw [blue, line width=1.2pt] (2,7) +(330:2cm) arc (330:360:2cm);
\draw [blue, line width=1.2pt] (2,7) +(0:2cm) arc (0:15:2cm);
\draw [blue, line width=1.2pt] (2,7) +(165:2cm) arc (165:210:2cm);
\draw [red, line width=1.2pt] (2,7) +(210:2cm) arc (210:320:2cm);


\draw[line width=1.2pt] (4,2) arc (360:180:2cm);
\draw[line width=1.2pt] (4,2) arc (0:15:2cm);
\draw[line width=1.2pt] (0,2) arc (180:165:2cm);

\draw [line width=1.2pt] (2,2) +(45:2cm) arc (45:135:2cm);
\draw [line width=1.2pt, dashed] (0.55,3.4)--(3.95,2.5);
\draw [blue, line width=1.2pt, dashed] (0.05,2.5)--(3.5,3.4);
\draw [blue, line width=1.2pt] (2,2) +(45:2cm) arc (45:90:2cm);
\draw [blue, line width=1.2pt] (2,2) +(165:2cm) arc (165:210:2cm);
\draw [red, line width=1.2pt] (2,2) +(210:2cm) arc (210:320:2cm);

\draw[line width=1.2pt] (12,7) arc (360:190:2cm);
\draw[line width=1.2pt] (12,7) arc (0:110:2cm);
\draw [gray, line width=1.2pt] (10,7) +(140:2.2cm) arc (140:165:2.2cm);
\draw [blue, line width=1.2pt] (10,7) +(330:2cm) arc (330:360:2cm);
\draw [blue, line width=1.2pt] (10,7) +(0:2cm) arc (0:90:2cm);

\draw [red, line width=1.2pt] (10,7) +(210:2cm) arc (210:320:2cm);


\draw[line width=1.2pt] (12,2) arc (360:190:2cm);
\draw[line width=1.2pt] (12,2) arc (0:110:2cm);
\draw [line width=1.2pt] (10,2) +(140:2.2cm) arc (140:165:2.2cm);
\draw [blue, line width=1.2pt] (10,2) +(330:2cm) arc (330:360:2cm);
\draw [blue, line width=1.2pt] (10,2) +(0:2cm) arc (0:90:2cm);

\draw [red, line width=1.2pt] (10,2) +(210:2cm) arc (210:320:2cm);

\node at (-1,0.5) {\footnotesize \textbf{Case A2}};
\node at (7,0.5) {\footnotesize \textbf{Case B2}};
\node at (-1,5.5) {\footnotesize \textbf{Case A1}};
\node at (7,5.5) {\footnotesize \textbf{Case B1}};

\path (0.3,1) coordinate (t2);
\path (8.3,1) coordinate (t02);

\path (2,4) coordinate (t3);
\path (10,4) coordinate (t03);

\path (3.6,0.8) coordinate (t4);
\path (11.6,0.8) coordinate (t04);

\path (0.3,6) coordinate (t20);
\path (8.3,6) coordinate (t020);

\path (2,9) coordinate (t30);
\path (10,9) coordinate (t030);

\path (3.6,5.8) coordinate (t40);
\path (11.6,5.8) coordinate (t040);

\path (0.52,3.41) coordinate (t5);
\path (3.48,3.41) coordinate (t05);
\path (0.52,8.41) coordinate (t50);
\path (3.48,8.41) coordinate (t050);

\path (0.05,2.5) coordinate (t6);
\path (3.95,2.5) coordinate (t06);
\path (0.05,7.5) coordinate (t60);
\path (3.95,7.5) coordinate (t060);

\path (9.3,3.85) coordinate (t7);
\path (8.3,3.4) coordinate (t07);
\path (7.9,2.6) coordinate (t8);
\path (8.04,1.7) coordinate (t08);

\path (9.3,8.85) coordinate (t70);
\path (8.3,8.4) coordinate (t070);
\path (7.9,7.6) coordinate (t80);
\path (8.04,6.7) coordinate (t080);
\draw [line width=1.2pt, dashed] (t70) .. controls (9,7.5) .. (t080);
\draw [gray, line width=1.2pt, dashed] (t070) .. controls (8.4,7.9) .. (t80);
\draw [line width=1.2pt, dashed] (t7) .. controls (9,3) .. (t8);
\draw [line width=1.2pt, dashed] (t07) .. controls (8.5,2.2) .. (t08);
\filldraw[black] (t2) circle(.2);
\filldraw[black] (t02) circle(.2);

\filldraw[black] (t3) circle(.2);
\filldraw[black] (t03) circle(.2);

\filldraw[black] (t4) circle(.2);
\filldraw[black] (t04) circle(.2);

\filldraw[black] (t20) circle(.2);
\filldraw[black] (t020) circle(.2);

\filldraw[black] (t30) circle(.2);
\filldraw[black] (t030) circle(.2);

\filldraw[black] (t40) circle(.2);
\filldraw[black] (t040) circle(.2);

\filldraw[black] (t5) circle(.08);
\filldraw[blue] (t05) circle(.08);
\filldraw[black] (t50) circle(.08);
\filldraw[black] (t050) circle(.08);

\filldraw[blue] (t6) circle(.08);
\filldraw[black] (t06) circle(.08);
\filldraw[blue] (t60) circle(.08);
\filldraw[blue] (t060) circle(.08);

\filldraw[black] (t7) circle(.08);
\filldraw[black] (t07) circle(.08);
\filldraw[black] (t8) circle(.08);
\filldraw[black] (t08) circle(.08);
\filldraw[black] (t70) circle(.08);
\filldraw[gray] (t070) circle(.08);
\filldraw[gray] (t80) circle(.08);
\filldraw[black] (t080) circle(.08);

\draw (t5) node[anchor=east] {\large $s$};
\draw (t05) node[anchor=west] {\large $t$};
\draw (t50) node[anchor=east] {\large $s$};
\draw (t050) node[anchor=west] {\large $t$};

\draw (t6) node[anchor=east] {\large $r$};
\draw (t06) node[anchor=west] {\large $u$};
\draw (t60) node[anchor=east] {\large $r$};
\draw (t060) node[anchor=west] {\large $u$};

\draw (t7) node[anchor=south] {\large $u$};
\draw (t07) node[anchor=east] {\large $t$};
\draw (t8) node[anchor=east] {\large $s$};
\draw (t08) node[anchor=east] {\large $r$};
\draw (t70) node[anchor=south] {\large $u$};
\draw (t070) node[anchor=east] {\large $t$};
\draw (t80) node[anchor=east] {\large $s$};
\draw (t080) node[anchor=east] {\large $r$};

\end{tikzpicture}
\caption{Two cases of replacement of arc-pairs (Cases A and B in Figure~\ref{fig:replicadeposito_opt})}\label{fig:remendadeposito_opt}
\end{center}
\end{figure}
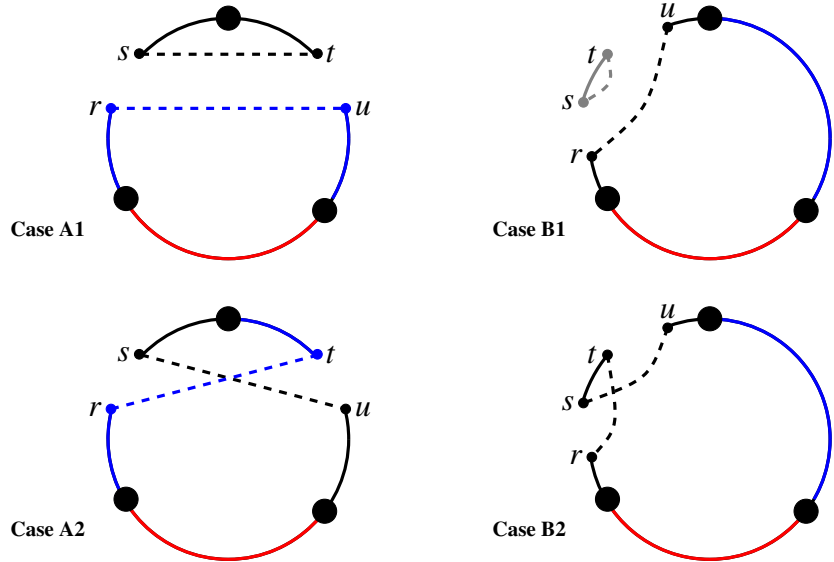

\begin{enumerate}
\item[]\noindent \textbf{STEP 1.} Let $\rho$ be the smallest number of nodes per route, excluding the base node. Transform the set of $m$ routes into 
a single route by replacing the base node by $m$ artificial copies of the base node (see Figure~\ref{fig:replicadeposito_opt}). Make a list of all possible pairs of distinct arcs in this single route.

\item[]\noindent \textbf{STEP 2.} In this step, the arc-pairs in the list are considered sequentially, until either a better solution is reached and 
return to Step 1 or there is no pair left, continue with Step 3. Let $\{\{r,s\},\{t,u\}\}$ be the current pair. Consider the two 
possibilities of replacement: (i) $\{\{r,t\},\{s,u\}\}$, (ii) $\{\{r,u\},\{s,t\}\}$. The replacement may split the single route into two 
routes (see Figure~\ref{fig:remendadeposito_opt}). If this happens, the new routes correspond to a feasible solution only if each of them contains at 
least one copy of the base node. Furthermore, the number of nodes in the route(s) between each pair of copies of the base node (or in the case where one of the subroutes created 
by the replacement contains only one copy of the base node) must be at least $\rho$. For each feasible solution, calculate its objective value. 
If there is an improvement, recover $m$ routes from the best improved single (or pair of) route(s) and go back to Step 1. If the end of the list is 
reached without (feasible) improvement, recover $m$ routes from the best improved single (or pair of) route(s) and go to Step 3.

\item[]\noindent \textbf{STEP 3.} Make a list of pairs of nodes, each distinct from the base node, and belonging to different routes. Consider the 
pairs sequentially. For each pair, calculate the value of the alternative solution obtained by swapping the nodes. Keep the solution 
with best objective function value.
\end{enumerate}

In Step 1 of the Balanced 2-opt, we transform the set of $m$ routes into a single route (Figure~\ref{fig:replicadeposito_opt}) in the same manner as described in~\cite{Art:lenstra1975some}. Case A and Case B in Figure~\ref{fig:replicadeposito_opt} are two possible configurations of arc-pairs 
in distinct routes and in the same route, respectively. The choice of $\rho$ as the smallest number of nodes per route, and the fact that 
of the arc-pair exchanges in Step 2 produces routes with at least $\rho$ nodes guarantees the balance among routes.

The exchanges of arcs in Step 2, 2-opt moves, are particular cases of $k$-opt move algorithm (LK) developed by~\cite{Art:lin1973effective}. In Step 3 the 
swapping of nodes between different routes can be considered as a 4-opt move, since the two arcs that share each exchanged node are deleted and replaced 
by four new arcs. Although the extension and generalization LKH-2 algorithm developed by~\cite{Art:helsgaun2009general} greatly improves LK, we chose 
2-opt and 4-opt moves, since they have a reasonably good performance and the balance constraints can be easily checked. Several patterns resulting 
from $k$-opt moves are not suitable to our model, since they destroy the balance and create disconnected subroutes. 
Figure~\ref{fig:remendadeposito_opt} (as in~\cite{Art:hachicha2000heuristics}) illustrates some examples of configurations after the replacement of arc-pairs, where four cases are obtained according to 
the relative position of each arc-pair, namely the arc-pairs being either in distinct routes (Cases A1 and A2) or in the same route (Cases B1 and B2).

In Case A1, the two disconnected routes obtained are feasible once the balance constraints are satisfied. Note that this exchange is not allowed in the 
$k$-opt moves. In Case A2, one can produce unbalanced solutions depending on the number of nodes between the copy of the base and an arc-pair node. Case B1 
is infeasible since one chain does not contain a copy of the base. Finally, in Case B2, there is a change in the order of nodes in a route, which may 
improve the solution.

\subsection{Multicover Elimination}

If a node in $W$ is covered by more than one node, we may improve the solution by eliminating one of the superfluous nodes. At this point, 
the initial set of $m$ routes constructed in Sector Partition has already undergone improvement in Phase 2. Note that only nodes in $V\setminus T$ (the support of $y$ that are not in $T$) may be removed from the solution. 
If a node $i$ may be removed, we consider the alternative solution obtained by 
deleting the arcs incident to $i$ and adding the arc linking its neighbors. Observe that, since the distances between nodes satisfy the triangle 
inequality, the removal of a node always implies in the decrease of the objective function.

The possibility of removal of node $i$ is checked 
in the brute force way: we delete it from the route and check whether this results in some node in $W$ being uncovered. This is a 
two-step procedure. In the first step we build a list of candidates for removal, and in the second step we examine the list sequentially, 
trying to remove as many nodes as possible.

\begin{enumerate}
\item[]\noindent \textbf{STEP 1.} Examine all nodes in the route that belong to $V\setminus T$ and build a list of superfluous nodes in descending 
order with respect to the amount of decrease in the objective function implied by their removal.

\item[]\noindent \textbf{STEP 2.} Consider the nodes in the list sequentially and remove the nodes whose removal 
does not destroy the required coverage and balance among routes.
\end{enumerate}

Note that this procedure is simpler than the Balanced 2-opt routine and it was chosen as the post-optimization for Sector Partition heuristic 
in order to reduce the computational time in Phase 3.

\section{Numerical experiments}\label{sec:Numerical}

The heuristics were tested on a set of adapted instances from TSPLIB and on an instance constructed from official data of the city of 
Vinhedo, S\~ao Paulo. They were implemented in $\textsc{Matlab}^{\scriptsize{\textregistered}}$ and tests were run on a PC with an Intel i7-4790S 3.20 GHz, with 8GB of RAM DDR3 and 
Linux operating system. The characteristics of the adapted instances are described next, followed by comments about the 
collection of data from the city of Vinhedo.

We close this section with statistics summarizing the test results and comments on the performances of the different heuristics.\\

\noindent{\it Data Instances}

The full characterization of an instance involves the following parameters: the cardinalities of the sets $V$, $T$, and $W$; the distance $c$ calculated 
from the sets $S_j$, $j\in W$; the number $m$; and the parameter $r$ (the maximum difference allowed between the number of nodes in different routes).

In order to observe the behavior of the heuristics with respect to scale, we have designed five classes of problems, where each class 
is defined by the total of nodes in $V\cup W$: the first class has $|V|=50$, $|W|=50$ corresponding to KroA100, KroB100, KroC100, KroD100 and KroE100; the second class $|V|=50$, $|W|=100$ corresponding to KroA150 and KroB150; the third class $|V|=100$, $|W|=100$ corresponding to KroA200 and KroB200; the fourth class $|V|=100$, $|W|=218$ corresponding to lin318; and the fifth class $|V|=200$, $|W|=200$ corresponding to rd400. Each class is subdivided in three subclasses, according to the cardinality of $T$: the subclass 1 has $|T|=\lceil|V|/8\rceil$; the subclass 2 has $|T|=\lceil|V|/4\rceil$; and the last one has $|T|=\lceil|V|/2\rceil$.

Each instance is characterized by a collection of pairs $\{(x_i,y_i)\ |\ i= 1,\ldots,|V|+|W|\}$, where both $x_i$ and $y_i$ are interpreted as coordinates of points in the Euclidean plane. The base coordinates are chosen from the collection so that it is centralized. Pairs 
corresponding to the nodes in $T^*$ are chosen sequentially in this list, and are followed 
by the pairs corresponding to the nodes in $V\setminus T$. All other pairs correspond to nodes in $W$.

Next we choose the constant $c$ used in the definition of the neighborhood $S_j$ of node $j\in W$. That is, the nodes in $V$ which are close 
enough to $j$ in the sense that $j$ will be considered covered if the route contains a node in $S_j$. The following considerations 
guide our choice. We want $|S_j|\geq 2$, for all $j$, so $c$ must be greater than the maximum of the 
distances $c_{jh(j)}$, for all $j\in W$, where $h(j)$ is the index of the node in $V$ that is the second closest node to $j$. Furthermore, 
we want that every node in $V\setminus T$ covers some node in $W$, otherwise this node could be eliminated from consideration in a 
pre-processing run. Thus, we also want $c$ to be greater than $m_h$, the distance from node $h$ to $W$, for all $h\in V\setminus T$. The 
value of $c$ is thus the greatest of the two maximum distances.

\begin{figure}[t]  
\begin{center}
\includegraphics[height=8cm, viewport=1 1 430 375, clip]{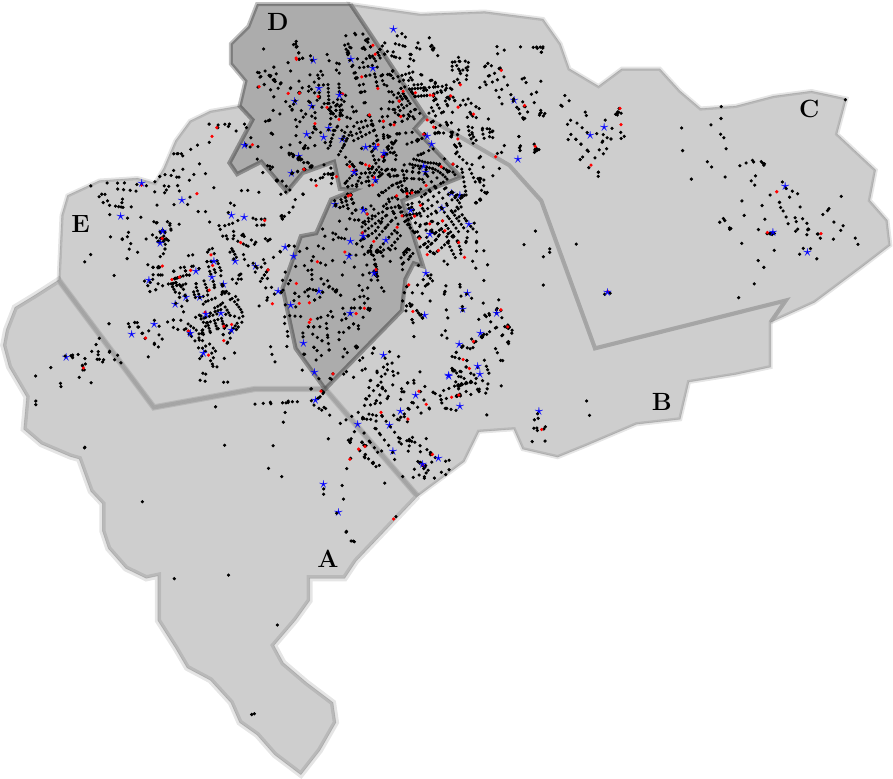}
\caption{Distribution of relevant nodes in the City of Vinhedo}\label{fig:DistributionVinhedo}
\end{center}
\end{figure}

The number of routes $m$ varied from 2 to 4 in the classes with 100 and 150 points, from 3 to 5 in the classes with 200 and 318 points, and from 4 to 6 in the class with 400 points. Classes with 100 and 150 points have $r=2$, the classes with 200 and 318 points have $r=3$, and the class with 400 points has $r=4$.\\

\noindent{\it City of Vinhedo Instance}

The Municipal Guard of the city of Vinhedo provided two lists of addresses. The first list, with 101 addresses, corresponded to sites that 
needed to be visited during rounds, and the second, with 133 addresses, contained the addresses of sites that should be covered. The geographical 
coordinates of these sites were calculated, as well as the coordinates of all street intersections. The geographical coordinates of the 
Municipal Guard base gave us the base node location. This real instance thus had 102 nodes in $T$, 133 nodes in $W$ and 2496 nodes 
in $V\setminus T$ (the intersections). Figure~\ref{fig:DistributionVinhedo} shows the distribution of points within the city region.

\begin{figure}[t]  
\begin{center}
\includegraphics[height=8cm, viewport=1 1 430 375, clip]{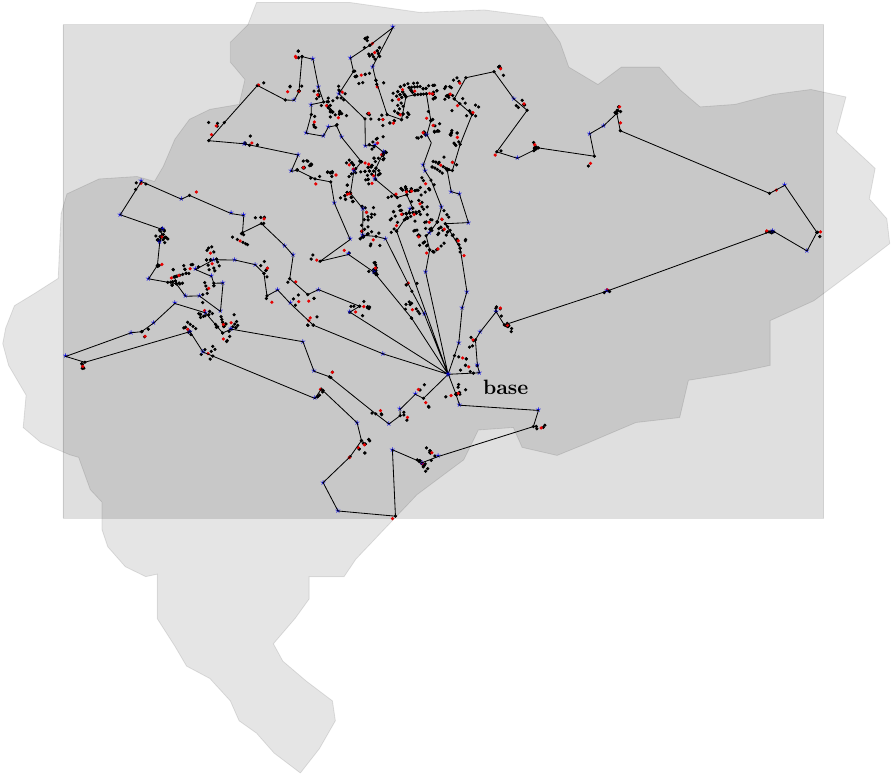}
\caption{Distribution of relevant nodes in the City of Vinhedo}\label{fig:RotasVinhedo}
\end{center}
\end{figure}

Empirical experience in security acquired by the Municipal Guard leads to the conclusion that a patrolling officer can watch each site up to a 
distance of 150 m, so we chose $c=150$ m. This allowed the elimination of 1563 nodes in $V\setminus T$, whose distances to $W$ were 
greater than $c$. We verify this by comparing the Figures~\ref{fig:DistributionVinhedo} and~\ref{fig:RotasVinhedo}. 
The latter illustrates a set of routes obtained in the tests. Note in Figure~\ref{fig:DistributionVinhedo} that the distribution of nodes in this instance is not uniform, 
for example, in Sector D there is a higher concentration of nodes than in Sector A. Our real-data instance had $|T|=102$, $|V\setminus T|=933$ and $|W|=133$. The values considered for $m$ were five, six and seven, 
which are related to the likely number of cars available for community patrolling duties. The parameter $r$ was chosen as six in the first case ($m=5$) and as 
eight in the remaining cases. According to the Municipal Guard this choice does not overload the patrolling officers. \\

\noindent{\it Summary of Results for TSPLIB Instances}

\begin{figure}[t]  
\begin{center}
\begin{tikzpicture}
\path (0.2,1.3) coordinate (p1);
\path (0.2,7.5) coordinate (p2);
\path (10,7.5) coordinate (p3);
\path (10,1.3) coordinate (p4);

\path (4.8598,4.5144) coordinate (base);
\path (8.8,3.9) coordinate (t1);
\path (2.3114,6.2984) coordinate (t2);
\path (6.0684,5.7514) coordinate (t3);
\path (4.8598,4.5144) coordinate (t4);
\path (9.1,5.625) coordinate (t5);

\path (7.6210,4.6216) coordinate (t6);
\path (4.5647,2.3938) coordinate (t7);
\path (3,4) coordinate (t8);
\path (8.2141,2.3884) coordinate (t9);
\path (4.4470,3.7386) coordinate (t10);
\path (6.1543,5.0898) coordinate (t11);
\path (7.7194,2.1) coordinate (t12);

\draw (5.53,4.515) node {\scriptsize \textbf{base}};

\draw[line width=.6pt] (p1) -- (p2) -- (p3) -- (p4) -- (p1);

\path (9.2181,5.8695) coordinate (t13);
\path (7.3821,3.9765) coordinate (t14);
\path (1.7627,2.5929) coordinate (t15);
\path (6.0571,4.9192) coordinate (t16);
\path (9.3547,3.8260) coordinate (t17);
\path (8.9,3.4) coordinate (t18);
\path (5.2,4.1) coordinate (t19);
\path (8.9365,4.3492) coordinate (t20);
\path (0.789,5.3) coordinate (t21);
\path (3.5287,6.6979) coordinate (t22);
\path (7.9,3.581) coordinate (t23);
\path (1.1,2.1) coordinate (t24);
\path (4.2,6.05) coordinate (t25);
\path (2.0277,6.6582) coordinate (t26);
\path (1.9872,1.9001) coordinate (t27);
\path (8.5,6.6) coordinate (t28);
\path (2.7219,6.1302) coordinate (t29);
\path (1,4.65) coordinate (t30);

\path (4.3510,6.573) coordinate (t33);
\path (9.3181,6.8940) coordinate (t34);
\path (4.6599,2.0911) coordinate (t35);
\path (4.4,4.6301) coordinate (t36);
\path (1.68,5.1) coordinate (t37);
\path (4.9,2.7) coordinate (t38);

\draw[line width=0.035cm] (t13) circle(.07);
\draw[line width=0.035cm] (t14) circle(.07);
\draw[line width=0.035cm] (t15) circle(.07);
\draw[line width=0.035cm] (t16) circle(.07);
\draw[line width=0.035cm] (t17) circle(.07);
\draw[line width=0.035cm] (t18) circle(.07);
\draw[line width=0.035cm] (t19) circle(.07);
\draw[line width=0.035cm] (t20) circle(.07);
\draw[line width=0.035cm] (t21) circle(.07);
\draw[line width=0.035cm] (t22) circle(.07);
\draw[line width=0.035cm] (t23) circle(.07);
\draw[line width=0.035cm] (t24) circle(.07);
\draw[line width=0.035cm] (t25) circle(.07);
\draw[line width=0.035cm] (t26) circle(.07);
\draw[line width=0.035cm] (t27) circle(.07);
\draw[line width=0.035cm] (t28) circle(.07);
\draw[line width=0.035cm] (t29) circle(.07);
\draw[line width=0.035cm] (t30) circle(.07);

\draw[line width=0.035cm] (t33) circle(.07);
\draw[line width=0.035cm] (t34) circle(.07);
\draw[line width=0.035cm] (t35) circle(.07);
\draw[line width=0.035cm] (t36) circle(.07);
\draw[line width=0.035cm] (t37) circle(.07);
\draw[line width=0.035cm] (t38) circle(.07);

\filldraw[very nearly transparent] (t13) circle(.6);
\filldraw[very nearly transparent] (t14) circle(.6);
\filldraw[very nearly transparent] (t15) circle(.6);
\filldraw[very nearly transparent] (t16) circle(.6);
\filldraw[very nearly transparent] (t17) circle(.6);
\filldraw[very nearly transparent] (t18) circle(.6);
\filldraw[very nearly transparent] (t19) circle(.6);
\filldraw[very nearly transparent] (t20) circle(.6);
\filldraw[very nearly transparent] (t21) circle(.6);
\filldraw[very nearly transparent] (t22) circle(.6);
\filldraw[very nearly transparent] (t23) circle(.6);
\filldraw[very nearly transparent] (t24) circle(.6);
\filldraw[very nearly transparent] (t25) circle(.6);
\filldraw[very nearly transparent] (t26) circle(.6);
\filldraw[very nearly transparent] (t27) circle(.6);
\filldraw[very nearly transparent] (t28) circle(.6);
\filldraw[very nearly transparent] (t29) circle(.6);
\filldraw[very nearly transparent] (t30) circle(.6);

\filldraw[very nearly transparent] (t33) circle(.6);
\filldraw[very nearly transparent] (t34) circle(.6);
\filldraw[very nearly transparent] (t35) circle(.6);
\filldraw[very nearly transparent] (t36) circle(.6);
\filldraw[very nearly transparent] (t37) circle(.6);
\filldraw[very nearly transparent] (t38) circle(.6);

\path (6.7214,6.9183) coordinate (t40);
\path (8.8,3.9) coordinate (t41);
\path (2,2.2) coordinate (t42);
\path (7.8,3.8086) coordinate (t43);
\path (3.7948,2.3377) coordinate (t44);
\path (8.4180,3.5) coordinate (t45);
\path (5.0281,1.5816) coordinate (t46);
\path (7.6,3.5) coordinate (t47);
\path (4.2889,5.3226) coordinate (t48);
\path (3.0462,3.7088) coordinate (t49);
\path (1.1,5.1) coordinate (t50);
\path (8.8,6.95) coordinate (t51);
\path (6.8222,2.6587) coordinate (t52);
\path (3.9,6.33) coordinate (t53);
\path (5.4167,4.7659) coordinate (t54);
\path (1.6087,2.2277) coordinate (t55);
\path (7.54,3.9748) coordinate (t56);
\path (3.7837,5.5595) coordinate (t57);
\path (8.9,6.6) coordinate (t58);
\path (9.12,3.9) coordinate (t59);

\filldraw[fill=black] (t41) circle(.06);
\filldraw[fill=black] (t42) circle(.06);
\filldraw[fill=black] (t43) circle(.06);

\filldraw[fill=black] (t45) circle(.06);

\filldraw[fill=black] (t47) circle(.06);

\filldraw[fill=black] (t50) circle(.06);
\filldraw[fill=black] (t51) circle(.06);

\filldraw[fill=black] (t53) circle(.06);

\filldraw[fill=black] (t55) circle(.06);
\filldraw[fill=black] (t56) circle(.06);

\filldraw[fill=black] (t58) circle(.06);
\filldraw[fill=black] (t59) circle(.06);

\draw (base)--(t8)--(t55)--(t50)--(t2)--(t53)--(base);
\draw (base)--(t3)--(t58)--(t5)--(t41)--(t6)--(t11)--(base);
\draw (base)--(t10)--(t7)--(t12)--(t9)--(t47)--(base);

\foreach \i in {2,...,12}
{
 \node at (t\i) {$\star$};
}
\end{tikzpicture}
\caption{Example of a feasible solution}\label{fig:RandomlyInstanceSolution}
\end{center}
\end{figure}
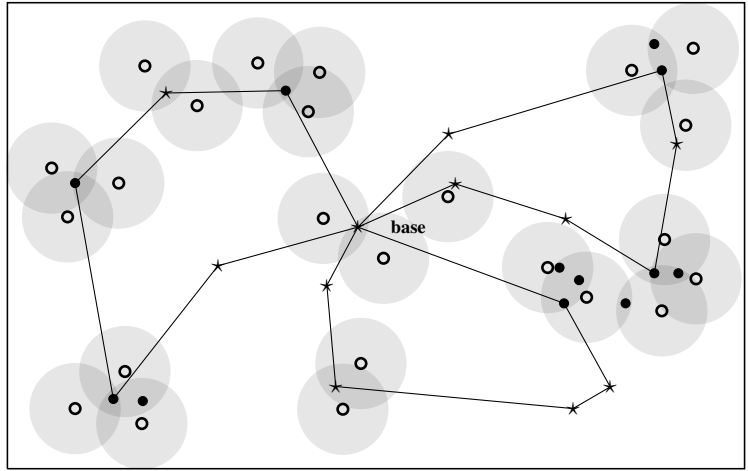

Figure~\ref{fig:RandomlyInstanceSolution} shows a typical solution of the problem. Notice that the three routes contain all nodes in $T$ (the star shaped nodes), but some nodes 
in $V\setminus T$ (bullets) were not picked. The gray disks are the neighborhoods of the nodes in $W$ (circles). One can verify that 
the solution is feasible with respect to coverage, as well as each gray disk contains at least one visited node. Furthermore, the 
routes are balanced, differing by at most one node.

\begin{table}[t]
{\scriptsize \begin{tabular}{cccccccccccccccccc}
\multicolumn{1}{l}{instance} & \multicolumn{1}{l|}{} & \multicolumn{ 4}{c|}{Greedy Selection} & \multicolumn{ 4}{c|}{Sector Partition} & \multicolumn{ 4}{c|}{Sweep Routine} & \multicolumn{ 4}{c}{Route 1$^{\mbox{\tiny st}}$/Cluster 2$^{\mbox{\tiny nd}}$} \\ \toprule
 & $m$ & $\tilde{r}$ & time  & cost & Q.I. & $\tilde{r}$ & time  & cost & Q.I. &$\tilde{r}$ & time  & cost & Q.I. & $\tilde{r}$ & time  & cost & Q.I. \\ \toprule
Kro-100 & 2 & 0 & 17.6 & 13465 & 1.377 & 1 & 7.1 & 10416 & 1.065 & 0 & 14.7 & 10498 & 1.074 & 0 & 9.4 & 9777 &\cellcolor{gray!30} 1 \\ 
A & 3 & 0 & 3.8 & 14962 & 1.160 & 2 & 3.2 & 12896 &\cellcolor{gray!30} 1 & 1 & 2.7 & 13194 & 1.023 & 0 & 3.0 & 13459 & 1.044 \\ 
 & 4 & 1 & 1.5 & 16045 & 1.078 & 1 & 2.5 & 14887 &\cellcolor{gray!30} 1 & 0 & 1.2 & 15186 & 1.020 & 1 & 2.8 & 15018 & 1.009 \\ &&&&&&&&&&&&&&&&&\\
 & 2 & 2 & 7.9 & 8827 &\cellcolor{gray!30} 1 & 2 & 3.9 & 8827 &\cellcolor{gray!30} 1 & 2 & 9.3 & 8827 &\cellcolor{gray!30} 1 & 0 & 3.0 & 10032 & 1.137 \\
B & 3 & 2 & 2.0 & 13176 & 1.197 & 1 & 2.8 & 11009 &\cellcolor{gray!30} 1 & 2 & 3.1 & 13183 & 1.197 & 1 & 1.6 & 12847 & 1.167 \\ 
 & 4 & 0 & 0.9 & 13526 & 1.027 & 1 & 1.7 & 13173 &\cellcolor{gray!30} 1 & 0 & 1.2 & 13610 & 1.033 & 1 & 1.2 & 13173 &\cellcolor{gray!30} 1 \\ &&&&&&&&&&&&&&&&&\\
 & 2 & 1 & 2.5 & 10606 & 1.087 & 1 & 2.6 & 9759 &\cellcolor{gray!30} 1 & 1 & 1.7 & 10606 & 1.087 & 1 & 2.1 & 9759 &\cellcolor{gray!30} 1 \\ 
C & 3 & 1 & 0.9 & 12095 & 1.074 & 2 & 1.4 & 11260 &\cellcolor{gray!30} 1 & 1 & 0.7 & 12450 & 1.106 & 0 & 1.0 & 12120 & 1.076 \\ 
 & 4 & 1 & 0.5 & 13640 & 1.060 & 2 & 1.2 & 12866 &\cellcolor{gray!30} 1 & 1 & 0.6 & 13699 & 1.065 & 1 & 1.1 & 13020 & 1.012 \\ &&&&&&&&&&&&&&&&&\\
 & 2 & 1 & 14.0 & 10526 & 1.022 & 1 & 9.6 & 10298 &\cellcolor{gray!30} 1 & 1 & 14.2 & 10526 & 1.022 & 1 & 8.1 & 10298 &\cellcolor{gray!30} 1 \\ 
D & 3 & 1 & 2.8 & 13306 & 1.045 & 2 & 5.0 & 12739 &\cellcolor{gray!30} 1 & 1 & 3.7 & 13279 & 1.042 & 0 & 2.4 & 13753 & 1.080 \\ 
 & 4 & 1 & 1.1 & 14777 & 1.034 & 1 & 3.2 & 14289 &\cellcolor{gray!30} 1 & 1 & 1.7 & 14474 & 1.013 & 1 & 1.8 & 14510 & 1.015 \\ &&&&&&&&&&&&&&&&&\\
 & 2 & 0 & 14.9 & 12191 & 1.150 & 0 & 3.8 & 10636 & 1.003 & 0 & 11.7 & 11534 & 1.088 & 0 & 5.8 & 10605 &\cellcolor{gray!30} 1 \\ 
E & 3 & 1 & 3.2 & 14510 & 1.051 & 2 & 2.5 & 13809 &\cellcolor{gray!30} 1 & 1 & 2.8 & 14504 & 1.050 & 2 & 2.2 & 14331 & 1.038 \\ 
 & 4 & 1 & 1.2 & 16392 & 1.061 & 2 & 1.7 & 15790 & 1.022 & 1 & 1.3 & 16787 & 1.086 & 2 & 1.8 & 15453 &\cellcolor{gray!30} 1 \\ \midrule
Kro-150 & 2 & 1 & 15.6 & 10275 &\cellcolor{gray!30} 1 & 2 & 10.5 & 10606 & 1.032 & 1 & 15.4 & 10275 &\cellcolor{gray!30} 1 & 1 & 11.9 & 11336 & 1.103 \\ 
A & 3 & 1 & 4.0 & 13385 & 1.030 & 2 & 4.3 & 13496 & 1.039 & 2 & 4.7 & 12992 &\cellcolor{gray!30} 1 & 1 & 4.6 & 13828 & 1.064 \\ 
 & 4 & 1 & 2.2 & 14790 & 1.030 & 2 & 4.5 & 14853 & 1.035 & 2 & 2.4 & 14357 &\cellcolor{gray!30} 1 & 1 & 2.9 & 14890 & 1.037 \\ &&&&&&&&&&&&&&&&&\\
 & 2 & 1 & 14.1 & 10395 & 1.043 & 1 & 10.1 & 10083 & 1.012 & 1 & 9.6 & 10083 & 1.012 & 0 & 4.3 & 9964 &\cellcolor{gray!30} 1 \\ 
B & 3 & 1 & 3.9 & 12967 & 1.114 & 2 & 5.9 & 11642 & 1.000 & 1 & 2.8 & 12967 & 1.114 & 1 & 2.8 & 11637 &\cellcolor{gray!30} 1 \\ 
 & 4 & 0 & 2.0 & 13769 & 1.063 & 2 & 3.4 & 13149 & 1.015 & 1 & 1.7 & 13696 & 1.058 & 1 & 2.4 & 12951 &\cellcolor{gray!30} 1 \\ \midrule
 Kro-200 & 3 & 1 & 5.9 & 14962 & 1.038 & 2 & 11.5 & 14408 &\cellcolor{gray!30} 1 & 1 & 10.0 & 15327 & 1.064 & 1 & 13.2 & 14962 & 1.038 \\ 
A & 4 & 0 & 3.9 & 16646 & 1.002 & 2 & 15.0 & 16869 & 1.015 & 0 & 5.5 & 16616 &\cellcolor{gray!30} 1 & 0 & 9.3 & 17250 & 1.038 \\ 
 & 5 & 2 & 3.2 & 18212 & 1.030 & 2 & 8.6 & 17674 &\cellcolor{gray!30} 1 & 1 & 3.5 & 19738 & 1.117 & 1 & 8.0 & 18151 & 1.027 \\ &&&&&&&&&&&&&&&&&\\
 & 3 & 1 & 69.1 & 16648 & 1.041 & 1 & 95.3 & 15993 &\cellcolor{gray!30} 1 & 1 & 68.3 & 16016 & 1.001 & 2 & 66.8 & 16161 & 1.010 \\ 
B & 4 & 2 & 33.5 & 19148 & 1.096 & 2 & 91.9 & 17933 & 1.026 & 2 & 31.6 & 17933 & 1.026 & 2 & 28.0 & 17473 &\cellcolor{gray!30} 1 \\ 
 & 5 & 2 & 13.4 & 21192 & 1.047 & 2 & 59.1 & 20440 & 1.010 & 2 & 12.9 & 20827 & 1.029 & 2 & 20.9 & 20243 &\cellcolor{gray!30} 1 \\ \midrule
\multicolumn{1}{l}{} & 3 & 2 & 44.3 & 15293 & 1.039 &  &  &  &  & 2 & 27.6 & 14713 &\cellcolor{gray!30} 1 & 1 & 32.6 & 16630 & 1.130 \\ 
lin318 & 4 & 2 & 23.0 & 19229 & 1.030 &  &  &  &  & 2 & 17.5 & 18670 &\cellcolor{gray!30} 1 & 0 & 17.7 & 19185 & 1.028 \\ 
 & 5 & 2 & 13.5 & 22009 &\cellcolor{gray!30} 1 &  &  &  &  & 2 & 11.2 & 22976 & 1.044 & 1 & 13.3 & 23390 & 1.063 \\ \midrule
\multicolumn{1}{l}{} & 4 & 3 & 998 & 8405 & 1.158 &  &  &  &  & 4 & 967 & 7256 &\cellcolor{gray!30} 1 & 1 & 650 & 7274 & 1.002 \\ 
rd400 & 5 & 3 & 754 & 8916 & 1.103 &  &  &  &  & 3 & 433 & 8084 &\cellcolor{gray!30} 1 & 2 & 340 & 8232 & 1.018 \\ 
 & 6 & 2 & 495 & 9311 & 1.094 &  &  &  &  & 4 & 270 & 8761 & 1.029 & 3 & 243 & 8512 &\cellcolor{gray!30} 1 \\ \bottomrule
\end{tabular}}
\caption{Results for subclass $|T|=\lceil|V|/8\rceil$}
\label{tab:Classe1}
\end{table}

Tables~\ref{tab:Classe1},~\ref{tab:Classe2} and~\ref{tab:Classe3} summarize the relevant data collected in the numerical experiments with the four heuristics listed in Table~\ref{tab:RoutePhase}. The first column indicates the instance, the second column the number of vehicle, and for each heuristic we report the $\tilde{r}$ (the maximum difference obtained between the number of nodes of different routes in the approximate solution), the time (in seconds) spent on each instance, the cost (total length of all tours) and the quality index (Q.I.) of the heuristic. This last number is the cost divided by the minimum cost over all four heuristics, showing how good the performance of that heuristic was, as compared to the one with the best cost. Therefore, the Q.I. of the routine with the best cost is 1, and the table cell containing this entry is shaded for emphasis.

\begin{table}[t]
{\scriptsize\begin{tabular}{cccccccccccccccccc}
\multicolumn{1}{l}{instance} & \multicolumn{1}{l|}{} & \multicolumn{ 4}{c|}{Greedy Selection} & \multicolumn{ 4}{c|}{Sector Partition} & \multicolumn{ 4}{c|}{Sweep Routine} & \multicolumn{ 4}{c}{Route 1$^{\mbox{\tiny st}}$/Cluster 2$^{\mbox{\tiny nd}}$} \\ \toprule
 & $m$ & $\tilde{r}$ & time  & cost & Q.I. & $\tilde{r}$ & time  & cost & Q.I. & $\tilde{r}$ & time  & cost & Q.I. & $\tilde{r}$ & time  & cost & Q.I. \\ \toprule
Kro-100 & 2 & 2 & 13.0 & 11781 & 1.076 & 2 & 12.7 & 10951 &\cellcolor{gray!30} 1 & 2 & 12.1 & 10951 &\cellcolor{gray!30} 1 & 1 & 9.0 & 11409 & 1.042 \\ 
A & 3 & 1 & 2.5 & 14943 & 1.195 & 2 & 6.8 & 12505 &\cellcolor{gray!30} 1 & 1 & 2.2 & 14216 & 1.137 & 1 & 3.0 & 13733 & 1.098 \\
 & 4 & 1 & 1.4 & 15026 &\cellcolor{gray!30} 1 & 1 & 5.0 & 15336 & 1.021 & 1 & 1.3 & 15203 & 1.012 & 1 & 1.5 & 15250 & 1.015 \\  &&&&&&&&&&&&&&&&&\\
 & 2 & 1 & 15.9 & 10975 &\cellcolor{gray!30} 1 & 1 & 33.5 & 11102 & 1.012 & 1 & 16.5 & 12340 & 1.124 & 1 & 15.0 & 10975 &\cellcolor{gray!30} 1 \\ 
B & 3 & 0 & 3.3 & 13245 &\cellcolor{gray!30} 1 & 2 & 3.1 & 13416 & 1.013 & 0 & 3.9 & 13245 &\cellcolor{gray!30} 1 & 0 & 5.8 & 13768 & 1.039 \\ 
 & 4 & 1 & 1.7 & 16358 & 1.029 & 2 & 1.8 & 16161 & 1.017 & 1 & 2.4 & 16511 & 1.039 & 1 & 2.8 & 15897 &\cellcolor{gray!30} 1 \\  &&&&&&&&&&&&&&&&&\\
 & 2 & 2 & 4.2 & 12982 &\cellcolor{gray!30} 1 & 2 & 25.5 & 12982 &\cellcolor{gray!30} 1 & 0 & 4.2 & 13224 & 1.019 & 2 & 4.1 & 12982 &\cellcolor{gray!30} 1 \\ 
C & 3 & 0 & 1.3 & 14211 & 1.012 & 1 & 12.0 & 14143 & 1.007 & 0 & 1.3 & 14043 &\cellcolor{gray!30} 1 & 0 & 1.4 & 14528 & 1.035 \\ 
 & 4 & 0 & 0.8 & 17227 & 1.100 & 2 & 8.5 & 15667 &\cellcolor{gray!30} 1 & 0 & 0.8 & 17328 & 1.106 & 0 & 0.8 & 17227 & 1.100 \\  &&&&&&&&&&&&&&&&&\\
 & 2 & 1 & 11.9 & 12739 & 1.101 & 1 & 18.3 & 11572 &\cellcolor{gray!30} 1 & 1 & 14.0 & 11855 & 1.024 & 1 & 12.8 & 11572 &\cellcolor{gray!30} 1 \\ 
D & 3 & 0 & 2.6 & 13522 &\cellcolor{gray!30} 1 & 2 & 2.6 & 13883 & 1.027 & 0 & 2.7 & 13522 &\cellcolor{gray!30} 1 & 0 & 3.4 & 13524 & 1.000 \\
 & 4 & 1 & 1.4 & 16552 & 1.075 & 2 & 8.2 & 15399 &\cellcolor{gray!30} 1 & 1 & 1.4 & 16033 & 1.041 & 1 & 2.8 & 15846 & 1.029 \\  &&&&&&&&&&&&&&&&&\\
 & 2 & 2 & 21.2 & 11700 & 1.093 & 2 & 15.5 & 11044 & 1.032 & 1 & 23.9 & 11029 & 1.031 & 1 & 16.8 & 10700 &\cellcolor{gray!30} 1 \\ 
E & 3 & 0 & 6.7 & 14252 & 1.072 & 2 & 8.3 & 13679 & 1.029 & 2 & 6.2 & 13296 &\cellcolor{gray!30} 1 & 0 & 3.4 & 14097 & 1.060 \\ 
 & 4 & 1 & 3.8 & 16947 & 1.061 & 2 & 6.3 & 15970 &\cellcolor{gray!30} 1 & 1 & 3.2 & 16855 & 1.055 & 1 & 2.7 & 16077 & 1.007 \\ \midrule
Kro-150 & 2 & 0 & 18.6 & 11413 & 1.030 & 1 & 14.1 & 11077 &\cellcolor{gray!30} 1 & 1 & 13.7 & 11077 &\cellcolor{gray!30} 1 & 1 & 9.1 & 11077 &\cellcolor{gray!30} 1 \\ 
A & 3 & 1 & 3.8 & 14333 & 1.064 & 2 & 6.8 & 13468 &\cellcolor{gray!30} 1 & 1 & 3.5 & 14113 & 1.048 & 1 & 3.2 & 13858 & 1.029 \\ 
 & 4 & 1 & 2.3 & 15239 & 1.039 & 2 & 6.2 & 15203 & 1.037 & 1 & 2.1 & 15166 & 1.034 & 1 & 2.3 & 14666 &\cellcolor{gray!30} 1 \\ &&&&&&&&&&&&&&&&&\\
 & 2 & 1 & 11.1 & 12685 & 1.032 & 2 & 17.5 & 12297 &\cellcolor{gray!30} 1 & 1 & 10.9 & 12384 & 1.007 & 1 & 10.3 & 12384 & 1.007 \\ 
B & 3 & 1 & 4.0 & 14170 & 1.003 & 1 & 5.4 & 14133 &\cellcolor{gray!30} 1 & 1 & 3.3 & 14843 & 1.050 & 1 & 3.7 & 14567 & 1.031 \\ 
 & 4 & 1 & 2.2 & 17697 & 1.125 & 2 & 6.1 & 15735 &\cellcolor{gray!30} 1 & 1 & 2.3 & 17186 & 1.092 & 1 & 2.9 & 15908 & 1.011 \\ \midrule
Kro-200 & 3 & 1 & 46.6 & 18033 & 1.061 & 3 & 88.9 & 16992 &\cellcolor{gray!30} 1 & 1 & 40.4 & 17187 & 1.011 & 1 & 45.0 & 17166 & 1.010 \\ 
A & 4 & 1 & 25.7 & 19745 & 1.026 & 3 & 103 & 19319 & 1.004 & 1 & 21.6 & 19246 &\cellcolor{gray!30} 1 & 1 & 26.3 & 19579 & 1.017 \\ 
 & 5 & 1 & 13.1 & 22718 & 1.038 & 3 & 14.1 & 22645 & 1.035 & 1 & 9.8 & 21888 &\cellcolor{gray!30} 1 & 1 & 14.9 & 21920 & 1.001 \\ &&&&&&&&&&&&&&&&&\\
\multicolumn{1}{l}{} & 3 & 1 & 199 & 17489 & 1.024 & 3 & 166 & 17818 & 1.043 & 1 & 158 & 17184 & 1.006 & 0 & 105 & 17084 &\cellcolor{gray!30} 1 \\ 
B & 4 & 2 & 100 & 19142 & 1.043 & 3 & 118 & 18813 & 1.025 & 2 & 83.1 & 19017 & 1.036 & 1 & 50.8 & 18357 &\cellcolor{gray!30} 1 \\ 
 & 5 & 2 & 49.0 & 21647 & 1.018 & 3 & 43.6 & 21705 & 1.021 & 2 & 37.2 & 21266 &\cellcolor{gray!30} 1 & 0 & 23.8 & 21787 & 1.024 \\ \midrule
\multicolumn{1}{l}{} & 3 & 2 & 268 & 16821 &\cellcolor{gray!30} 1 &  &  &  &  & 2 & 202 & 16866 & 1.003 & 1 & 126 & 16975 & 1.009 \\ 
lin318 & 4 & 2 & 126 & 20804 & 1.013 &  &  &  &  & 2 & 95.0 & 20810 & 1.013 & 1 & 114 & 20541 &\cellcolor{gray!30} 1 \\ 
 & 5 & 2 & 56.8 & 23984 & 1.011 &  &  &  &  & 2 & 39.6 & 23727 &\cellcolor{gray!30} 1 & 1 & 33.5 & 23996 & 1.011 \\ \midrule
\multicolumn{1}{l}{} & 4 & 2 & 1243 & 9338 & 1.184 &  &  &  &  & 3 & 670 & 7886 &\cellcolor{gray!30} 1 & 1 & 959 & 8334 & 1.057 \\ 
rd400 & 5 & 2 & 760 & 9602 & 1.140 &  &  &  &  & 3 & 410 & 8424 &\cellcolor{gray!30} 1 & 2 & 678 & 8736 & 1.037 \\ 
 & 6 & 1 & 510 & 10631 & 1.150 &  &  &  &  & 2 & 285 & 9246 &\cellcolor{gray!30} 1 & 1 & 345 & 9301 & 1.006 \\ \bottomrule
\end{tabular}}
\caption{Results for subclass $|T|=\lceil|V|/4\rceil$}
\label{tab:Classe2}
\end{table}

\begin{table}[t]
{\scriptsize \begin{tabular}{cccccccccccccccccc}
\multicolumn{1}{l}{instance} & \multicolumn{1}{l|}{} & \multicolumn{ 4}{c|}{Greedy Selection} & \multicolumn{ 4}{c|}{Sector Partition} & \multicolumn{ 4}{c|}{Sweep Routine} & \multicolumn{ 4}{c}{Route 1$^{\mbox{\tiny st}}$/Cluster 2$^{\mbox{\tiny nd}}$} \\ \toprule
 & $m$ & $\tilde{r}$ & time  & cost & Q.I. & $\tilde{r}$ & time  & cost & Q.I. & $\tilde{r}$ & time  & cost & Q.I. & $\tilde{r}$ & time  & cost & Q.I. \\ \toprule
Kro-100 & 2 & 2 & 122 & 13062 &\cellcolor{gray!30} 1 & 2 & 196 & 13150 & 1.007 & 0 & 123 & 13216 & 1.012 & 0 & 123 & 13216 & 1.012 \\ 
A & 3 & 0 & 39.3 & 15966 & 1.079 & 2 & 146 & 14801 &\cellcolor{gray!30} 1 & 0 & 39.9 & 15410 & 1.041 & 0 & 38.6 & 16628 & 1.123 \\ 
 & 4 & 0 & 7.8 & 17937 &\cellcolor{gray!30} 1 & 2 & 11.2 & 17980 & 1.002 & 0 & 7.9 & 17940 & 1.000 & 0 & 8.3 & 18107 & 1.009 \\ &&&&&&&&&&&&&&&&&\\
 & 2 & 2 & 116 & 14052 & 1.114 & 2 & 127 & 12609 &\cellcolor{gray!30} 1 & 0 & 118 & 14075 & 1.116 & 0 & 117 & 14111 & 1.119 \\ 
B & 3 & 0 & 38.4 & 18619 & 1.113 & 2 & 35.7 & 17497 & 1.046 & 0 & 38.9 & 16903 & 1.011 & 0 & 39.7 & 16727 &\cellcolor{gray!30} 1 \\ 
 & 4 & 0 & 7.6 & 19165 & 1.023 & 2 & 6.7 & 18879 & 1.007 & 0 & 7.4 & 18744 & 1.000 & 0 & 7.8 & 18742 &\cellcolor{gray!30} 1 \\ &&&&&&&&&&&&&&&&&\\
 & 2 & 0 & 128 & 15287 &\cellcolor{gray!30} 1 & 2 & 134 & 15395 & 1.007 & 0 & 130 & 15287 &\cellcolor{gray!30} 1 & 0 & 130 & 15287 &\cellcolor{gray!30} 1 \\ 
C & 3 & 0 & 40.4 & 17157 & 1.012 & 2 & 166 & 16961 &\cellcolor{gray!30} 1 & 0 & 41.7 & 17034 & 1.004 & 0 & 40.4 & 17292 & 1.020 \\ 
 & 4 & 0 & 7.6 & 19577 & 1.001 & 2 & 8.1 & 19570 & 1.001 & 0 & 7.6 & 19731 & 1.009 & 0 & 7.9 & 19556 &\cellcolor{gray!30} 1 \\ &&&&&&&&&&&&&&&&&\\
 & 2 & 0 & 118 & 13854 & 1.046 & 2 & 145 & 13248 &\cellcolor{gray!30} 1 & 0 & 119 & 13680 & 1.033 & 0 & 116 & 13675 & 1.032 \\ 
D & 3 & 0 & 38.8 & 16375 & 1.012 & 1 & 41.2 & 16367 & 1.012 & 0 & 39.5 & 16178 &\cellcolor{gray!30} 1 & 0 & 39.4 & 16777 & 1.037 \\ 
 & 4 & 0 & 10.1 & 18158 & 1.012 & 2 & 121 & 18259 & 1.017 & 0 & 9.5 & 18442 & 1.028 & 0 & 9.3 & 17946 &\cellcolor{gray!30} 1 \\ &&&&&&&&&&&&&&&&&\\
 & 2 & 0 & 115 & 12542 &\cellcolor{gray!30} 1 & 0 & 102 & 12542 &\cellcolor{gray!30} 1 & 0 & 112 & 12542 &\cellcolor{gray!30} 1 & 0 & 111 & 12542 &\cellcolor{gray!30} 1 \\ 
E & 3 & 0 & 38.8 & 16837 & 1.030 & 1 & 36.3 & 16423 & 1.004 & 0 & 39.1 & 16352 &\cellcolor{gray!30} 1 & 0 & 38.7 & 16837 & 1.030 \\ 
 & 4 & 0 & 9.2 & 18229 & 1.038 & 2 & 70.8 & 17567 &\cellcolor{gray!30} 1 & 0 & 9.2 & 18086 & 1.030 & 0 & 9.4 & 17755 & 1.011 \\ \midrule
Kro-150 & 2 & 0 & 158 & 13881 & 1.005 & 0 & 171 & 15304 & 1.108 & 0 & 158 & 13881 & 1.005 & 0 & 158 & 13809 &\cellcolor{gray!30} 1 \\ 
A & 3 & 0 & 44.9 & 16366 & 1.090 & 2 & 68.1 & 15021 &\cellcolor{gray!30} 1 & 0 & 43.6 & 16481 & 1.097 & 0 & 43.5 & 15862 & 1.056 \\ 
 & 4 & 0 & 13.3 & 18795 & 1.028 & 1 & 15.4 & 18399 & 1.006 & 0 & 13.3 & 19389 & 1.061 & 0 & 13.5 & 18281 &\cellcolor{gray!30} 1 \\ &&&&&&&&&&&&&&&&&\\
 & 2 & 0 & 169 & 15096 &\cellcolor{gray!30} 1 & 2 & 191 & 15181 & 1.006 & 0 & 169 & 15236 & 1.009 & 0 & 169 & 15096 &\cellcolor{gray!30} 1 \\ 
B & 3 & 0 & 44.5 & 17211 & 1.015 & 0 & 75.4 & 16951 &\cellcolor{gray!30} 1 & 0 & 43.4 & 16951 &\cellcolor{gray!30} 1 & 0 & 43.8 & 16951 &\cellcolor{gray!30} 1 \\ 
 & 4 & 0 & 12.3 & 19582 & 1.018 & 2 & 18.4 & 19349 & 1.006 & 0 & 12.0 & 19341 & 1.006 & 0 & 12.6 & 19230 &\cellcolor{gray!30} 1 \\ \midrule
Kro-200 & 3 & 1 & 702 & 22610 & 1.166 & 3 & 980 & 19385 &\cellcolor{gray!30} 1 & 1 & 778 & 19513 & 1.007 & 1 & 751 & 19836 & 1.023 \\
A & 4 & 1 & 349 & 23658 & 1.074 & 3 & 767 & 22220 & 1.009 & 1 & 375 & 22019 &\cellcolor{gray!30} 1 & 1 & 375 & 22232 & 1.010 \\ 
 & 5 & 0 & 182 & 27281 & 1.130 & 3 & 608 & 24437 & 1.012 & 0 & 194 & 24415 & 1.011 & 0 & 188 & 24152 &\cellcolor{gray!30} 1 \\ &&&&&&&&&&&&&&&&&\\
 & 3 & 1 & 720 & 23552 & 1.151 & 3 & 957 & 20946 & 1.023 & 1 & 755 & 20471 &\cellcolor{gray!30} 1 & 1 & 762 & 20534 & 1.003 \\ 
B & 4 & 1 & 361 & 25711 & 1.186 & 3 & 761 & 22032 & 1.016 & 1 & 379 & 21683 &\cellcolor{gray!30} 1 & 1 & 379 & 22042 & 1.017 \\ 
 & 5 & 0 & 186 & 27893 & 1.155 & 3 & 553 & 24703 & 1.023 & 0 & 195 & 24572 & 1.018 & 0 & 195 & 24149 &\cellcolor{gray!30} 1 \\ \midrule
\multicolumn{1}{l}{} & 3 & 1 & 1158 & 17724 & 1.004 &  &  &  &  & 1 & 1170 & 17660 &\cellcolor{gray!30} 1 & 1 & 856 & 18088 & 1.024 \\ 
lin318 & 4 & 2 & 688 & 21101 & 1.018 &  &  &  &  & 2 & 700 & 21311 & 1.028 & 1 & 466 & 20724 &\cellcolor{gray!30} 1 \\ 
 & 5 & 2 & 436 & 23740 &\cellcolor{gray!30} 1 &  &  &  &  & 2 & 458 & 24398 & 1.028 & 1 & 330 & 24083 & 1.014 \\ \midrule
\multicolumn{1}{l}{} & 4 & 1 & 2610 & 10207 & 1.103 &  &  &  &  & 1 & 2630 & 9252 &\cellcolor{gray!30} 1 & 1 & 2641 & 9321 & 1.007 \\ 
rd400 & 5 & 1 & 1568 & 11501 & 1.160 &  &  &  &  & 1 & 1603 & 9917 &\cellcolor{gray!30} 1 & 1 & 1572 & 10345 & 1.043 \\ 
 & 6 & 1 & 1098 & 11612 & 1.091 &  &  &  &  & 1 & 1077 & 10645 &\cellcolor{gray!30} 1 & 1 & 1103 & 11256 & 1.057 \\ \bottomrule
\end{tabular}}
\caption{Results for subclass $|T|=\lceil|V|/2\rceil$}
\label{tab:Classe3}
\end{table}

For instances up to 150 points, Sector Partition has the biggest number of the Q.I. equals to 1, approximate 71\% of total, followed by Route-first/Cluster-second with 60\% of total. The worst Q.I. is 1.377 (37.7\% worse than the best result), achieved by the Greedy Selection heuristic, and the average value of Q.I. is 1.033. We conclude that the performances of the various heuristics with respect to the quality aspect were very similar.

For remaining instances, Sweep Routine has the biggest number of the Q.I. equals to 1, approximate 53\% of total. Since the distribution of points in the instances with 318 and 400 points is non-uniform, the Sector Partition could not produce balanced routes in those instances. And the average value of Q.I. is 1.032, which again indicates similar performances with respect to the cost.

\begin{figure}[t]
\begin{center}
\includegraphics[height=9cm, viewport=25 10 410 300, clip]{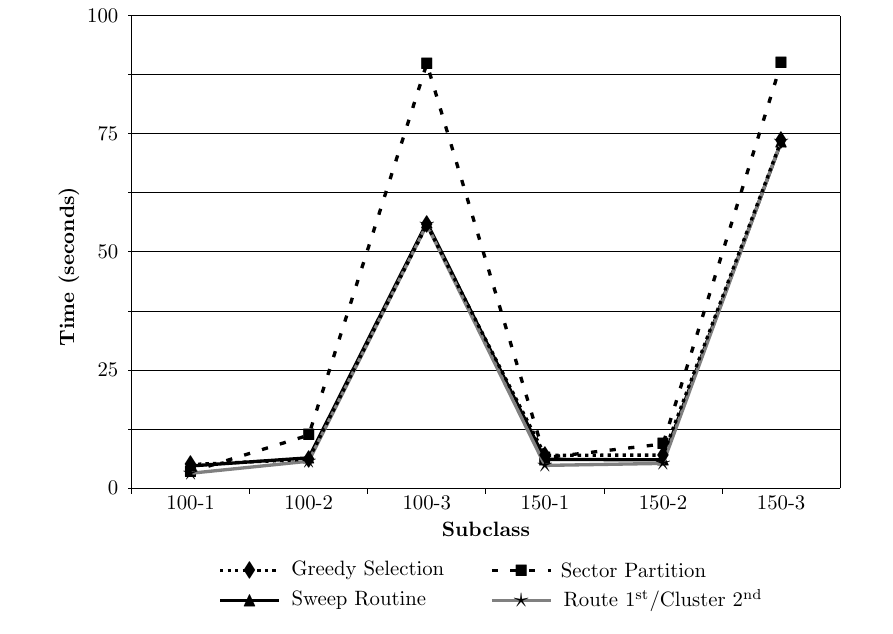}
\caption{Average times for subclasses 100-1 to 150-3} \label{fig:GraficoResults1}
\end{center}
\end{figure}

\begin{figure}[t]
\begin{center}
\centerline{\includegraphics[height=9cm, viewport=20 10 415 300, clip]{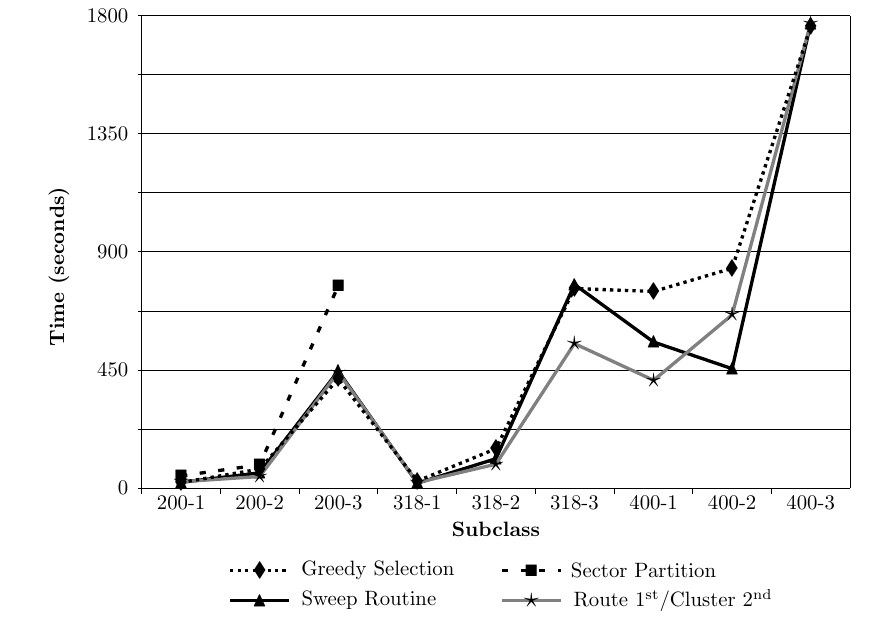}} 
\caption{Average times for subclasses 200-1 to 400-3} \label{fig:GraficoResults2}
\end{center}
\end{figure}

We observed discrepancies in the computational effort (measured by the average time spent) of the heuristics, as illustrated by the 
charts in Figures~\ref{fig:GraficoResults1} and~\ref{fig:GraficoResults2}. The disparity of ranges led to the construction of two distinct charts. Figure~\ref{fig:GraficoResults1} contains data relative to subclasses 100-1 to 150-3, the data of the remaining subclasses are in Figure~\ref{fig:GraficoResults2}.

In the first set of subclasses, the Route-first/Cluster-second is the fastest, except for the subclass 150-3, where Sweep Routine does slightly better. This pattern is maintained as the size increases, with the exception of class 400-2, where its time is 45\% bigger than the minimum time.

This analysis shows that the Sector Partition is well-adapted to smaller instances, where points are uniformly distributed. For larger instances the Sweep Routine obtained solutions with better cost. In terms of speed, the Route-first/Cluster-second obtained better computation times overall, but the performances of all heuristics were very similar in this aspect.

Next, we make a comparison between our results and the ones obtained in~\cite{Art:Ha}, which considers the closest problem variation found in the literature. The main differences between the formulations are that in theirs, the number of routes is not fixed, there is a constraint for the number of visits allowed on each route, and the routes may not be balanced.

In order to make a fair comparison, we consider instances in which the number of vehicles in the solution is the same in both approaches, and the maximum number of visits in our solution does not exceed their value.

The results are summarized in Table~\ref{tab:comparativo}, which is divided into two parts, the first part contains the results from literature, namely the name of the instance, the cardinality of $T$, the maximum number $p$ of points allowed on each route, the number of vehicles, the computational time and the cost. The second part contains the our results for the same instances from TSPLIB, which satisfy the proposed conditions for the comparison. This part contains the name of the instance, $|T|$, the maximum number $\tilde{p}$ of points obtained on each route, $m$, the value of $\tilde{r}$ (the maximum difference obtained between the number of nodes of different routes in the solution), the heuristic considered, time, cost, and the ratio of the cost of our heuristic to theirs.

Analyzing the table we can notice that most of the costs obtained in our solutions were very close to the ones obtained in~\cite{Art:Ha}, and the ratio of all instances varied between 0.828 and 1.185. However, it is important to emphasize that even though we are comparing the same instances from TSPLIB, it was not possible to reproduce the choices of $T$, $V\setminus T$ and $W$, since only the number of points in each set was provided in their paper. Another point worth noting is that there is no guarantee that their solutions are balanced. Finally, the proximity of the costs obtained in both formulations can be used to measure the quality of our heuristics.\\    

\begin{table}[t]
{\scriptsize \begin{tabular}{lcccccccccclccc}
\multicolumn{ 6}{c|}{\cite{Art:Ha}} & \multicolumn{ 9}{c}{Our method} \\ \toprule
instance & $|T|$ & $p$ & $m$ & time & cost & instance & $|T|$ & $\tilde{p}$ & $m$ & $\tilde{r}$ & heuristic & time  & cost & ratio \\ \toprule
A1-1-50-50-6 & 1 & 6 & 2 & 0.8 & 9130 &  & 7 & 6 & 2 & 0 & Route/Cluster & 10.4 & 9777 & 1.071 \\ 
A1-10-50-50-5 & 10 & 5 & 4 & 3.8 & 15440 & KroA100 & 13 & 4 & 4 & 1 & Greedy Selection  & 1.4 & 15026 & 0.973 \\ 
A1-10-50-50-6 & 10 & 6 & 3 & 3.9 & 14064 &  & 13 & 6 & 3 & 2 & Sector Partition & 6.8 & 12505 & 0.889 \\ \midrule
B1-1-50-50-5 & 1 & 5 & 2 & 0.6 & 9723 &  & 7 & 5 & 2 & 0 & Route/Cluster & 3.0 & 10032 & 1.032 \\ 
B1-1-50-50-6 & 1 & 6 & 2 & 0.6 & 9382 &  & 7 & 6 & 2 & 2 & Sweep Routine & 9.3 & 8827 & 0.941 \\ 
B1-10-50-50-4 & 10 & 4 & 4 & 2.5 & 15209 & KroB100 & 13 & 4 & 4 & 1 & Route/Cluster & 2.8 & 15897 & 1.045 \\ 
B1-10-50-50-5 & 10 & 5 & 3 & 2.1 & 13535 &  & 13 & 5 & 3 & 0 & Greedy Selection  & 3.3 & 13245 & 0.979 \\ 
B1-10-50-50-8 & 10 & 8 & 2 & 2.0 & 10344 &  & 13 & 8 & 2 & 1 & Greedy Selection  & 16 & 10975 & 1.061 \\ \midrule
C1-1-50-50-4 & 1 & 4 & 3 & 0.6 & 11372 &  & 7 & 4 & 3 & 2 &  & 1.4 & 11260 & 0.990 \\ 
C1-1-50-50-5 & 1 & 5 & 2 & 0.7 & 9900 & KroC100 & 7 & 5 & 2 & 1 & Sector Partition & 2.6 & 9759 & 0.986 \\ 
C1-10-50-50-4 & 10 & 4 & 4 & 2.2 & 18212 &  & 13 & 4 & 4 & 2 &  & 8.5 & 15667 & 0.860 \\ \midrule
D1-1-50-50-4 & 1 & 4 & 3 & 0.9 & 11606 &  & 7 & 4 & 3 & 0 & Route/Cluster & 2.4 & 13753 & 1.185 \\ 
D1-1-50-50-8 & 1 & 8 & 2 & 0.9 & 9361 & \multirow{2}*{KroD100} & 7 & 7 & 2 & 1 & Sector Partition & 9.6 & 10298 & 1.100 \\ 
D1-10-50-50-5 & 10 & 5 & 4 & 3.4 & 18576 &  & 13 & 5 & 4 & 2 & Sector Partition & 8.2 & 15399 & 0.829 \\ 
D1-10-50-50-6 & 10 & 6 & 3 & 2.9 & 16330 &  & 13 & 5 & 3 & 0 & Greedy Selection  & 2.6 & 13522 & 0.828 \\ \midrule
A2-20-100-100-6 & 20 & 6 & 5 & 40 & 20966 & \multirow{2}*{KroA200} & 25 & 6 & 5 & 1 & Sweep Routine & 9.8 & 21888 & 1.044 \\ 
A2-20-100-100-8 & 20 & 8 & 4 & 42 & 18418 &  & 25 & 8 & 4 & 3 & Sector Partition & 103 & 19319 & 1.049 \\ \midrule
B2-20-100-100-8 & 20 & 8 & 5 & 114 & 22156 & KroB200 & 25 & 7 & 5 & 2 & Sweep Routine & 37 & 21266 & 0.960 \\ \bottomrule
\end{tabular}}
\caption{Comparison with method of~\cite{Art:Ha}}
\label{tab:comparativo}
\end{table}

\noindent{\it Vinhedo Instance: Comparison of Heuristics}

In practice, a patrolling car is assigned to a geographical area, and the patrolling officers know the locations of the visits and the points that must 
be observed in this area. More precisely, due to the non-uniform distribution of the geographical points, whenever the number of available patrolling 
cars is five, two of them are assigned to Sectors A, B and C, two cars are assigned to Sector D, and one car to Sector E. If six cars are available, one car 
is assigned to each of the sectors A, B, C and E, and two cars to Sector D. Finally, if seven cars are available, one car 
is assigned to each of the sectors A, B, C and E, and three cars to Sector D. After interviewing the patrolling officers, we noticed that given a 
geographical sector and set of points to be visited, the officers generally choose their routes according to a greedy rule (Empirical routes), where the next point to be 
visited is the nearest from the current one, which may be unproductive.

Table~\ref{tab:ResultsVinhedo} reports the cost (in kilometers), quality index (Q.I.), time (in seconds), 
the value chosen for the parameter $r$, the value $\tilde{r}$ obtained, the size of the fleet, and the heuristic considered together with the empirical routes used by the patrolling officers. The numerical experimental data are collected in three cases: fleet of size five, six or seven.

\begin{table}[t]
\begin{center}
\begin{tabular}{rcccccl} \toprule 
\multicolumn{1}{c}{Cost} &\multirow{2}*{Q.I.} &\multirow{1}*{Time} &\multirow{2}*{$r$}  & \multirow{2}*{$\tilde{r}$} &\multirow{2}*{\footnotesize Vehicles}    &\multirow{2}*{Heuristic}  \\ 
\multicolumn{1}{c}{(Km)}&  & (sec)  &  &  &  &  \\ \toprule
106.271& 1.12099 & 55024 & 6 & 3 & 5  & Greedy Selection \\ 
94.801&\cellcolor{gray!30} 1 & 50590 & 6 & 2 & 5  & Sweep Routine \\ 
\multirow{1}*{99.486}& \multirow{1}*{1.04942} & \multirow{1}*{41759} & \multirow{1}*{6} & \multirow{1}*{3} & \multirow{1}*{5} &  {\small Route 1$^{\mbox{\scriptsize st}}$/Cluster 2$^{\mbox{\scriptsize nd}}$} \\ 
134.857& \multirow{1}*{1.42253} & --& \multirow{1}*{6} & 18 & \multirow{1}*{5} &  {Empirical}\\ \midrule
114.022& 1.08799 & 38887 & 8 & 4 & 6  & Greedy Selection \\
104.800&\cellcolor{gray!30} 1 & 36110 & 8 & 3 & 6  & Sweep Routine \\
\multirow{1}*{104.806}& \multirow{1}*{1.00006} & \multirow{1}*{27084} & \multirow{1}*{8} & \multirow{1}*{1} & \multirow{1}*{6} &  {\small Route 1$^{\mbox{\scriptsize st}}$/Cluster 2$^{\mbox{\scriptsize nd}}$} \\
140.735& \multirow{1}*{1.34289} & --& \multirow{1}*{8} & 35 & \multirow{1}*{6} &  {Empirical}\\ \midrule
121.494& 1.10100 & 25013 & 8 & 4 & 7  & Greedy Selection \\
110.349&\cellcolor{gray!30} 1 & 23913 & 8 & 4 & 7  & Sweep Routine \\
\multirow{1}*{111.234}& \multirow{1}*{1.00802} & \multirow{1}*{18670} & \multirow{1}*{8} & \multirow{1}*{2} & \multirow{1}*{7} &  {\small Route 1$^{\mbox{\scriptsize st}}$/Cluster 2$^{\mbox{\scriptsize nd}}$} \\ 
147.700& \multirow{1}*{1.33848} & --& \multirow{1}*{8} & 35 & \multirow{1}*{7} &  {Empirical}\\ \bottomrule
\end{tabular}\caption{Results from the Vinhedo instance}\label{tab:ResultsVinhedo}
\end{center}
\end{table}

The Sector Partition heuristic is omitted from Table~\ref{tab:ResultsVinhedo} because it did not obtain balanced feasible solutions, due to the extremely non-central location of the base of operations. Recall that in the adapted instances from TSPLIB, the base coordinates was chosen from the collection data so that the base was centralized. Thus, in this case, the partition produced by the heuristic obtained a set of balanced routes. 

Unfortunately, this did not happen in the Vinhedo instance. The geographical sites considered fit in a rectangular area of aspect ratio 1.5, that is, the 
width to height ratio (see Figure~\ref{fig:RotasVinhedo}, which shows the routes built by the Sweep Routine heuristic in the 5-vehicle case). If we were to 
consider the ``base region" adopted for the TSPLIB instances, the base of operations should be located in a rectangle of sides 30\% of the 
bigger rectangle, with same center. In the Vinhedo instance the base lies below and outside this ideal rectangle, missing the region 
by 6\% of the vertical dimension. 

Furthermore, the geographical distribution of sites that need to be visited or covered is highly non-uniform. 
If we divide the shaded rectangle of Figure~\ref{fig:RotasVinhedo} in four rectangles by drawing vertical and horizontal lines intersecting at the 
base, we will find 75.3\% of all nodes concentrated in the northwest rectangle, 15.2\% in the northeast, 6.8\% in the southwest and only 
2.7\% in the southeast rectangle. As a result, the Sector Partition heuristic produced quite unbalanced routes, with the maximum
difference obtained between two routes exceeding the value of the parameter $r$.

The speed exhibited by the Sector Partition heuristic suggests that it is worthwhile to further investigate its 
possibilities. One direction for future research is that of a non-uniform division in sectors, choosing the central angles in such a 
way that the number of nodes in each sector is approximately equal.

\begin{figure}[t]  
\begin{center}
\includegraphics[height=8cm, viewport=1 1 430 375, clip]{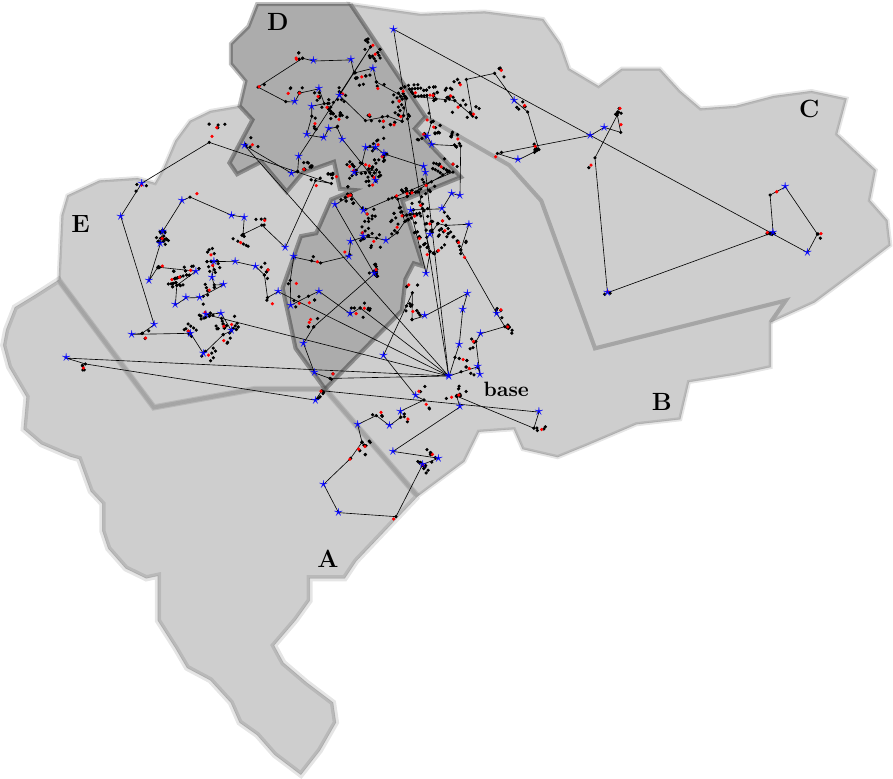}
\caption{Routes obtained empirically by patrolling officers}\label{fig:RoutesEmpirical}
\end{center}
\end{figure}

Compared with the artificial instances, the bad performance of the Greedy Selection heuristic was confirmed in the (very large scale) Vinhedo instance, with the highest Q.I. in 
every test considered. However, the Empirical routes obtained by the Vinhedo police were worse than any of the heuristics considered. 
Figure~\ref{fig:RoutesEmpirical} illustrates the solution for five cars, where two cars are 
responsible for Sectors A, B and C, two cars are assigned to Sector D, and one car is responsible for Sector E. Considering the overall length of the routes 
in the first two columns of Table~\ref{tab:ResultsVinhedo}, the Empirical routes were about 33\% to 42\% worse than the Sweep Routine 
(see a example of a solution for five vehicles in Figure~\ref{fig:RotasVinhedo}). When compared with the 
Greedy Selection, the Empirical routes were about 21\% to 27\% worse. 

In discordance with the previous tests, the Sweep Routine scored better objective function values, although not by much, than the 
Route-first/Cluster-second heuristic, in all cases considered, as indicated by the shaded cells in Table~\ref{tab:ResultsVinhedo}. However, the Sweep 
Routine exhibited 13\% to 25\% larger times than Route-first/Cluster-second.

Since the geographical points of interest are more concentrated in some sectors, the Empirical routes were very unbalanced, which can be seen in the fifth column of 
Table~\ref{tab:ResultsVinhedo}. Note that there are routes differing by 35 nodes. Moreover, in the case of five vehicles, Table~\ref{tab:SweepVSEmpirical} shows the 
length and number of visits for each individual route. There is a difference of up to 18 visits and the average route lengths 
in the Sweep Routine is about 30\% lower.

The solutions obtained were considered acceptable since a patrolling car may cover this distance several 
times during a work shift. This contemplates very well the objective of providing a good coverage with the available resources.	

\begin{table}[!htb]
\begin{center}
\begin{tabular}{ccccccc}
\toprule
&&\multicolumn{2}{c}{Sweep Routine} & & \multicolumn{2}{c}{Empirical} \\ \cmidrule{3-4} \cmidrule{6-7}
&&\multicolumn{1}{c}{Cost (Km)} &  \multirow{1}*{ Visits} && \multicolumn{1}{c}{Cost (Km)}  &\multirow{1}*{Visits}\\ \toprule
Route 1& & 22.3154& 37& &29.3413 &33 \\
Route 2& & 22.9958& 39& &38.9261 &47 \\
Route 3& & 15.0461& 37& &15.9927 &29 \\
Route 4& & 17.3789& 39& &24.0204 &39 \\
Route 5& & 17.1356& 37& &26.5766 &40 \\ \midrule
Average& &18.9744& &&26.9714 &  \\\bottomrule
\end{tabular}\caption{Comparative between Sweep Routine and Empirical Routes}\label{tab:SweepVSEmpirical}
\end{center}
\end{table}

\section{Conclusions and future work}\label{sec:conclusions}

We developed, implemented and tested various heuristics for the construction of routes in the context of urban patrolling. They were 
compared in numerical tests involving a set of adapted instances from TSPLIB, and variants of a problem defined with real data from the city of Vinhedo, Brazil. 

We developed a modified Sweep Routine, which produced solutions with lower costs for larger instances, while the Sector Partition heuristic, developed by us produced solutions with lower costs for smaller instances (up to 150 points). The computational time was quite uniform for all heuristics.

The heuristics were tested using the Vinhedo instance. The Greedy Selection heuristic, when compared with the other heuristics, 
produced solutions with lower quality. The solutions of our modifications of the Sweep Routine and Route-first/Cluster-second Routine had similar quality, 
though the former was always better. On the other hand, the latter was consistently faster.

The solutions obtained in the Vinhedo instance show the advantage of using this kind of modeling and heuristics for large scale problems, since the numerical results show 
that the Empirical routes used by the Municipal Guard are not balanced, interfering with the contact between the patrolling officers and the members 
of the community, which is the main objective of the community policing. We conclude that the division in sectors used by the Guard can be 
improved by using our solutions (see Figures~\ref{fig:RoutesEmpirical} and~\ref{fig:RotasVinhedo}).

We observed that the Sector Partition heuristic needs improvement in order to be applied in real instances having highly non-uniform 
distribution of nodes, turning it into a competitive and robust routine. This will be the subject of future research.

\section*{Acknowledgements}

The first author was partially support by CAPES and FAEPEX (grant 534/09). The second author was partially supported by CNPq-PRONEX Optimization and FAPESP (grant 2006/53768-0). The authors thank Osmir Cruz, Commander of the Municipal Guard of Vinhedo, the municipal government of Vinhedo and the company SSR-Tecnologia for translating the list of addresses into a list of geographical coordinates.

\end{document}